\documentclass[runningheads]{llncs}

\usepackage{eccv}

\usepackage{graphicx} \graphicspath{{figures/}}
\usepackage{amsmath}
\usepackage{amssymb}
\usepackage{booktabs}

\definecolor{cvprblue}{rgb}{0.21,0.49,0.74}
\usepackage[pagebackref,breaklinks,colorlinks,citecolor=cvprblue]{hyperref}
\usepackage{enumitem}
\usepackage{multirow}
\usepackage{adjustbox}
\usepackage{colortbl} 
\usepackage[capitalize]{cleveref}
\crefname{section}{Sec.}{Secs.}
\Crefname{section}{Section}{Sections}
\Crefname{table}{Table}{Tables}
\crefname{table}{Tab.}{Tabs.}
\usepackage[dvipsnames]{xcolor}
\usepackage[misc]{ifsym}

\newcommand{\inc}[1]{\textcolor{black}{#1 $\uparrow$}}

\newcommand{\model}{\textsc{PQ3D}\xspace}

\usepackage{eccvabbrv}

\usepackage{graphicx}
\usepackage{booktabs}

\usepackage[accsupp]{axessibility}  

\usepackage{hyperref}

\usepackage{orcidlink}

\begin{document}

\title{Unifying 3D Vision-Language Understanding via Promptable Queries}

\author{
\begin{tabular}{cccccc}
Ziyu Zhu$^{1,2\dagger}$\thanks{Work done as an intern at BIGAI.~~~\Letter~Corresponding author. $\dagger$ BNRist, THUAI, Department of Computer Science, Tsinghua University, Beijing 100084, China.}\orcidlink{0000-0003-1556-0791}, & Zhuofan Zhang$^{1,2}$\orcidlink{0009-0005-9965-4745}, & Xiaojian Ma$^{2}$\orcidlink{0000-0001-5609-3822}, &  Xuesong Niu$^{2}$\orcidlink{0000-0001-7737-4287}, & Yixin Chen$^{2}$\orcidlink{0000-0002-8176-0241},   
\end{tabular}
\\
\begin{tabular}{cccc}
Baoxiong Jia$^{2}$\orcidlink{0000-0002-4968-3290}, & Zhidong Deng$^{1\text{\Letter}\dagger}$\orcidlink{0000-0001-9970-1023}, & Siyuan Huang$^{2\text{\Letter}}$\orcidlink{0000-0003-1524-7148}, & Qing Li$^{2\text{\Letter}}$\orcidlink{0000-0003-1185-5365}
\end{tabular}
}


\authorrunning{Z.~Zhu et al.}

\institute{\begin{tabular}{cc}
$^1$Tsinghua University & $^2$State Key Laboratory of General Artificial Intelligence, BIGAI
\end{tabular} \\
\textbf{\href{https://pq3d.github.io}{pq3d.github.io}}
}


\maketitle

\begin{abstract}
A unified model for 3D vision-language (3D-VL) understanding is expected to take various scene representations and perform a wide range of tasks in a 3D scene. However, a considerable gap exists between existing methods and such a unified model, due to the independent application of representation and insufficient exploration of 3D multi-task training. In this paper, we introduce \textbf{\model}, a unified model capable of using \textbf{\underline{P}}romptable \textbf{\underline{Q}}ueries to tackle a wide range of \textbf{\underline{3D}}-VL tasks, from low-level instance segmentation to high-level reasoning and planning. This is achieved through three key innovations: (1) unifying various 3D scene representations (\ie, voxels, point clouds, multi-view images) into a shared 3D coordinate space by segment-level grouping, (2) an attention-based query decoder for task-specific information retrieval guided by prompts, and (3) universal output heads for different tasks to support multi-task training. Tested across ten diverse 3D-VL datasets, \model demonstrates impressive performance on these tasks, setting new records on most benchmarks. Particularly, \model improves the state-of-the-art on ScanNet200 by 4.9\% (AP25), ScanRefer by 5.4\% (acc@0.5), Multi3DRefer by 11.7\% (F1@0.5), and Scan2Cap by 13.4\% (CIDEr@0.5). 
Moreover, \model supports flexible inference with individual or combined forms of available 3D representations, \eg, solely voxel input.
\end{abstract}

\section{Introduction}

Recent advancements in embodied artificial intelligence have emphasized the importance of connecting 3D scene understanding with natural language~\cite{embodied, embodiedlang, teach, 3d-vista,huang2023embodied}. This step is crucial for embodied agents to understand and execute human instructions in real-world scenarios~\cite{saycan, sayplan}. In recent years, numerous tasks and datasets for benchmarking 3D scene understanding with languages have been proposed, including 3D semantic segmentation~\cite{scannet200}, 3D vision-language (3D-VL) reasoning (referring~\cite{jia2024sceneverse,scanrefer,referit3d,multi3drefer,scanents}, question answering~\cite{scanqa,sqa3d,3dgqa}, and captioning~\cite{scan2cap}), and open-vocabulary 3D understanding~\cite{semab,takmaz2023openmask3d,kerr2023lerf,guo2024semantic}. 

The state-of-the-art (SOTA) approaches typically address these tasks utilizing specific scene representations~\cite{mask3d, transrefer3d, scanqa, 3dsps, vil3dref, viewrefer}. These representations can all be derived from RGB-D streams, each with unique advantages and drawbacks. For instance, voxels offer a uniform, grid-like structure ideal for instance segmentation~\cite{scannet200,mask3d} but struggle to capture fine geometric details. Point clouds provide detailed spatial information crucial for visual grounding~\cite{3d-vista,vil3dref}, but they miss texture details. Multi-view images include a rich visual context beneficial for open-vocabulary scene understanding~\cite{peng2023openscene,takmaz2023openmask3d} but lack accurate 3D location. The full potential of combining all these representations for holistic scene understanding has not been well studied. Additionally, several works in 3D-VL reasoning explored multi-task training~\cite{unit3d,3djcg} and pretraining ~\cite{3d-vista, 3d-contrastive} to seek mutual benefits from related tasks; however, they still require an off-the-shelf 3D mask proposal module. Thus, despite substantial progress, there is still a gap toward a \textbf{unified model} achieving comprehensive 3D-VL understanding, from low-level instance segmentation to high-level reasoning and planning.

\begin{figure}[t]
\centering
    \includegraphics[width=1.0\textwidth]{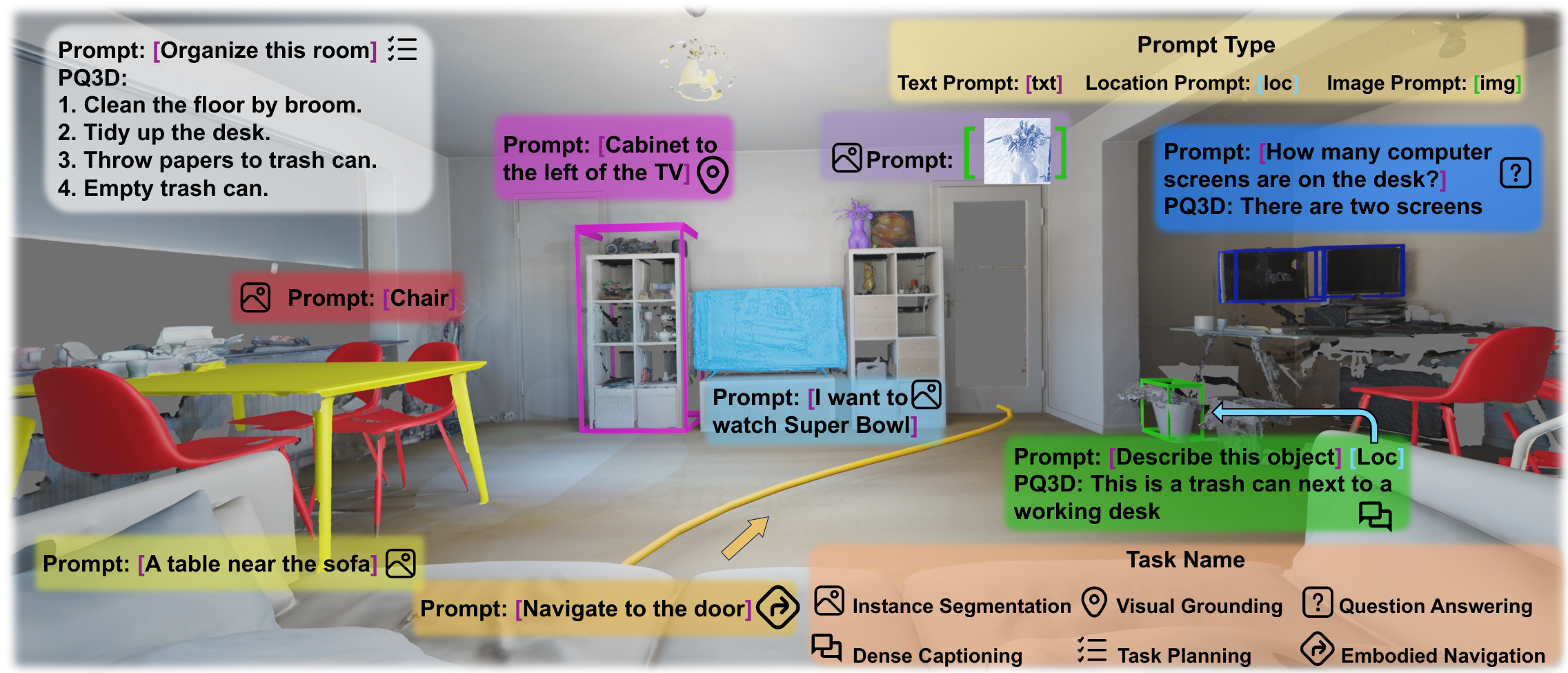}
    \captionof{figure}{\model is a unified model for 3D vision-language understanding, capable of taking various prompts (object categories, referring sentences, images, locations) to perform a wide range of tasks in a 3D scene, including instance segmentation, visual grounding, question answering and dense captioning. Remarkably, \model can take a novel prompt type unseen during training, \eg, an image sketch of a vase, to locate the related object in the scene.   If further instruction-tuned with a large language model and plugged into an embodied agent, \model can also
    plan a complex task and navigate the agent to desired objects.}
    \label{fig:teaser}
\end{figure}




We expect such a unified model for 3D scene understanding to integrate various scene representations and select appropriate features based on specific task instructions. However, two significant challenges exist: Firstly, these different scene representations have varying granularities, making it difficult to unify them in a single feature space. Secondly, there have been limited research efforts regarding developing a unified training paradigm that can accommodate 3D-VL tasks at different levels with diverse input instructions and output formats.

\begin{figure}[!t]
    \centering
    \includegraphics[width=1.0\linewidth]{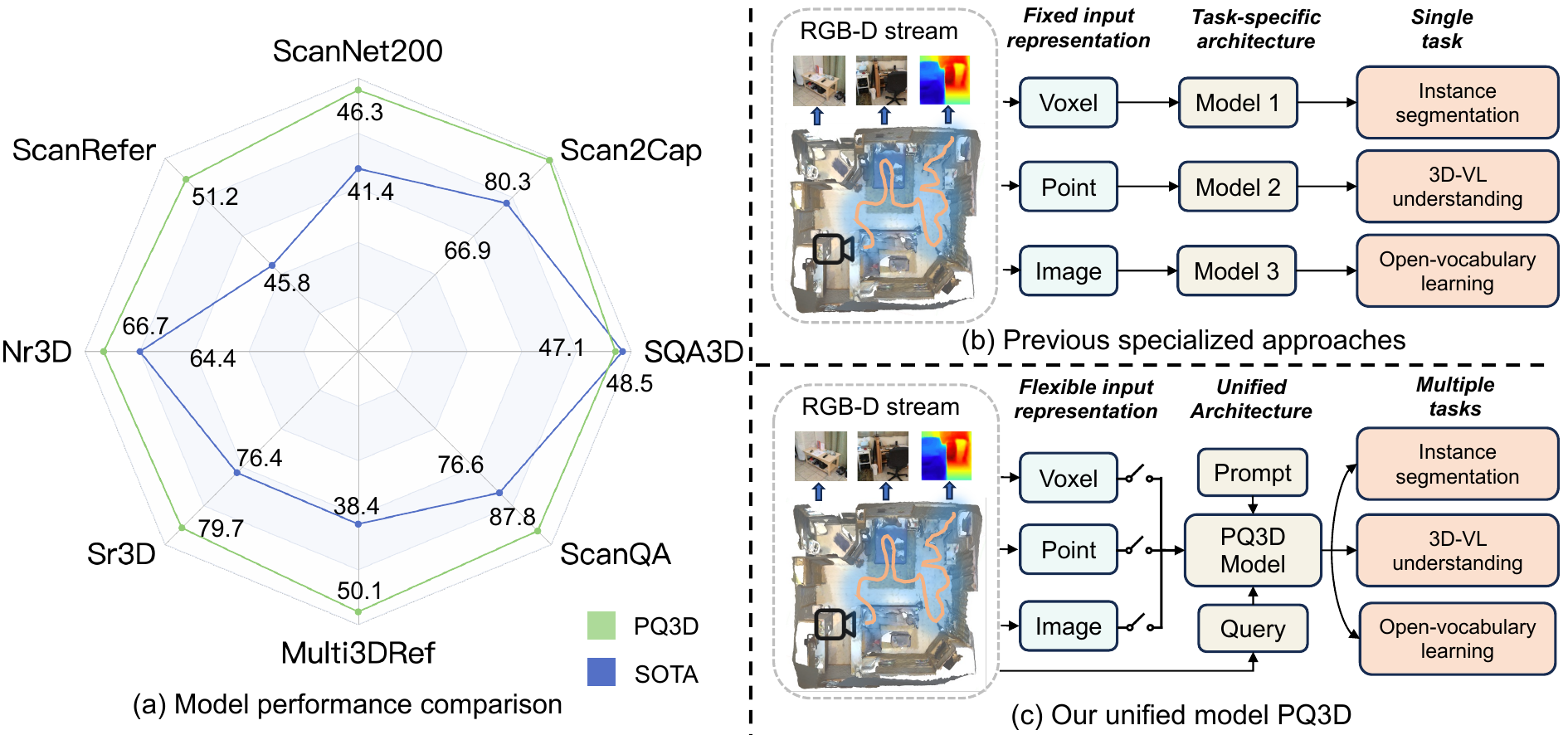}
    \caption{
    Comparison between \model and other models. (a) When comparing \model to other state-of-the-art (SOTA) methods, \model demonstrates superior performance on most tasks. (b)  Previous models have been designed for specific tasks and representations, often limiting the potential for developing a unified model. (c) Our \model can flexibly accommodate various input representations, effectively addressing a wide range of tasks.
    }
    \label{fig:radar}
\end{figure}

To address these challenges, we introduce \textbf{\model}, a unified model using \textbf{\underline{P}}romptable \textbf{\underline{Q}}ueries to concurrently manage various \textbf{\underline{3D}} scene \textit{representations}, \textit{prompts}, and \textit{outputs} in numerous 3D-VL tasks, as depicted in \cref{fig:teaser}. An overview of the proposed \model model is shown in \cref{fig:radar}(c), featuring three key innovations: (1) We unify dense point cloud features with multi-scale voxel features and multi-view image features into a shared 3D coordinate space. This process involves the unsupervised grouping of 3D points into larger segments and pooling features to the segment level, significantly reducing the number of points and facilitating training. (2) A novel attention-based query decoder is introduced that progressively retrieves task-specific information from aligned scene features under the guidance of task prompts. (3) Each query is processed through three universal output heads to predict an instance mask, a task-relevance score, and a sentence; these are then combined to produce the required task outputs.

We conduct extensive experiments on ten 3D-VL datasets, including ScanNet200~\cite{scannet200} / Replica~\cite{replica} for instance segmentation, ScanRefer~\cite{scanrefer} / ReferIt3D~\cite{referit3d} / Multi3DRef~\cite{multi3drefer} for visual grounding, ScanQA~\cite{scanqa} / SQA3D for question-answering, Scan2Cap~\cite{scan2cap} for dense captioning, and ObjNav from CortexBench~\cite{majumdar2023we} for embodied navigation. The proposed \model achieves impressive results across these tasks, setting new records on most tasks as shown in \cref{fig:radar}(a). For example, our model boosts the state-of-the-art on ScanNet200 by 4.9\% (AP25), ScanRefer by 5.4\% (acc@0.5), Multi3DRef by 11.7\% (F1@0.5), and Scan2Cap by 13.4\% (CIDEr@0.5). More importantly, \model is the first \textbf{unified model} capable of handling all these tasks simultaneously. The proposed model also shows zero-shot capability with novel prompt types; for instance, we can prompt it with an image sketch to locate all related objects in a scene as in \cref{fig:teaser}.

Our main contributions can be summarized as follows:
\begin{itemize}[leftmargin=*,noitemsep,topsep=0pt]
    \item We introduce \model, a unified model adept at solving a broad spectrum of 3D-VL tasks with promptable queries. The tasks range from low-level instance segmentation to visual grounding, and high-level reasoning and planning.
    \item Our model uniquely aligns voxels, point clouds, and multi-view images into a shared 3D space and employs an attention-based query decoder to adaptively extract task-relevant features guided by prompts, offering a flexible approach to model all 3D-VL tasks.
    \item In our extensive experimentation across various 3D-VL tasks, \model not only achieves competitive results but also sets new records in most of the tasks. 
\end{itemize}

\section{Related Work}
\subsection{3D Vision-language Learning}
In recent times, there has been a surging interest in the field of 3D vision-language (3D-VL) learning. 3D-VL tasks establish a vital connection between the physical world and natural language, contributing to the development of embodied intelligence~\cite{3d-vista,jia2024sceneverse}.
In this emerging domain, objects refer~\cite{scanrefer, referit3d, scanents, multi3drefer}, question answering~\cite{scanqa, 3dqa}, and dense captioning~\cite{scan2cap}  tasks are introduced to evaluate natural language grounding concerning 3D object properties and relationships. Additional tasks include embodied question answering~\cite{sqa3d} and navigation~\cite{habitat}, which explore models' capabilities within an embodied environment.
 
Numerous models have been proposed to tackle these benchmarks in task-specific ways. MVT~\cite{mvt3d}, ViL3DRel~\cite{vil3dref}, and ViewRefer~\cite{viewrefer} perform 3D visual grounding by explicitly incorporating spatial relation information into their designs. While these endeavors~\cite{3dvg, butd, 3dsps, sat, eda, vil3dref, viewrefer, mvt3d} have demonstrated impressive results, they often require specialized model architectures tailored to specific tasks. In contrast, our model is unified, performing multiple tasks in one model. In a recent line of research, several works~\cite{huang2023embodied, 3d-llm, ll3da} explore 3D instruction tuning to leverage the power of large language model (LLM) to enhance 3D reasoning. \model can further benefit these models by providing a robust 3D scene encoding module guided by language, which can be seamlessly integrated into these instruction tuning pipelines.

\subsection{Query-based Model}
Query-based models have gained prominence in 2D perception~\cite{detr, inst-as-queries, fastinst, dong2021solq, hu2021istr}. DETR~\cite{detr} has allowed queries to attend to visual features from the backbone network dynamically, enabling object detection without the need for post-processing techniques like NMS~\cite{detr,inst-as-queries}. Building upon this concept, subsequent works such as~\cite{inst-as-queries, fastinst, mask2former} have extended the query-based approach to universal image segmentations, where queries are employed to match pixel-level features and generate high-quality masks.  Additionally, the use of queries initialized from language has led to advancements in referring segmentation~\cite{language-as-queries, ding2021vision}.

More recent literature has further demonstrated the potential of query-based models in achieving unified understanding~\cite{perceiver-io, li2023blip, xdecoder, yang2022unified}.  For instance, Perceiver-IO~\cite{perceiver-io} leverages queries to extract information from multi-modal features, while XDecoder~\cite{xdecoder} formulates multiple vision-language tasks within a generalized decoding framework. However, in the context of 3D vision-language understanding, a query-based paradigm for handling multiple tasks is still lacking.

\subsection{Promptable Segmentation}
The concept of promptable segmentation, as presented in the SAM framework~\cite{SAM}, centers around the utilization of prompts to direct the process of segmentation. These prompts can manifest in diverse forms, such as foreground/background points, approximate bounding boxes or masks, and free-form text, which furnish guidance on the desired elements to be segmented in an image~\cite{SAM, liu2023simpleclick, lin2020interactive}. 

Promptable segmentation is intimately linked with open-vocabulary learning. Previous studies such as OWL-ViT~\cite{owl-vit} and OVR-CNN~\cite{ovrcnn} have leveraged contrastive learning with extensive image-text pairs to demonstrate object detection generalization capabilities~\cite{regionclip, owl-vit, ovrcnn}. OpenSeg~\cite{openseg} and OVSeg~\cite{ovseg} extend open-vocabulary detection from the object level to pixel-level segmentation~\cite{lseg, openseg, ovseg, xu2023learning, xu2023masqclip, qin2023freeseg}. In the field of 3D vision, methods like OpenMask3D~\cite{takmaz2023openmask3d} and OpenScene~\cite{peng2023openscene} have used 2D features to accomplish zero-shot segmentation. The PLA model~\cite{ding2023pla} harnessed captions to align vision and language for novel-class instance segmentation. Our work takes cues from these existing methodologies, leveraging the benefits of prompting images, text, and point-based data to achieve open-vocabulary promptable segmentation. 

\section{Method}

\begin{figure*}[!ht]
    \centering
    \includegraphics[width=\linewidth]{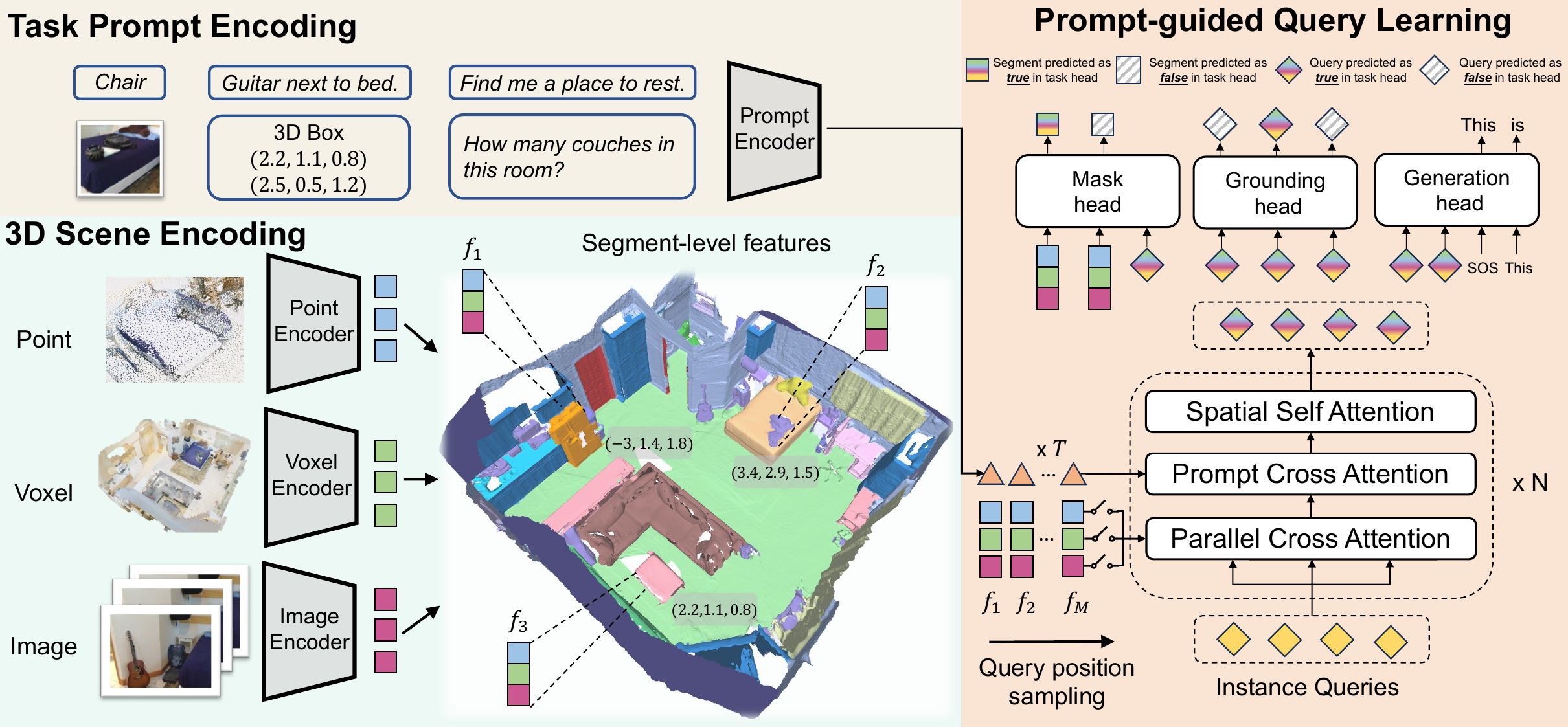}
    \caption{The model architecture of \model, which consists of \textbf{Task Prompt Encoding}, \textbf{3D Scene Encoding}, and \textbf{Prompt-guided Query Learning} modules. In prompt encoding, task prompts in diverse formats are projected to a shared feature space. In scene encoding, point clouds, voxel grids, and multi-view images of a scene are first encoded by corresponding encoders and then aligned into a shared 3D coordinate space. The prompt-guided query learning module takes in zero-initialized instance queries and progressively retrieves task-relevant information from aligned scene features under the guidance of task prompts. Finally, each updated instance query is fed into three output heads to predict an instance mask, a task-relevance score, and a sentence.}
    \label{fig:model}
\end{figure*}

In this section, we present \model, which consists of three main modules:  Task Prompt Encoding, 3D Scene Encoding, and Prompt-guided Query Learning, as depicted in \cref{fig:model}. Next, we will explain the details of each module.

\subsection{Task Prompt Encoding}
In various 3D-VL tasks, a task prompt can be of diverse formats, including object categories, referring sentences, questions, 3D bounding boxes, 3D locations, images, \etc. Diverse prompt formats are one of the key obstacles to building a unified model for 3D-VL understanding. However, we figured out that all these diverse prompts can be divided into three categories: \textit{textual}, \textit{visual}, and \textit{numerical}. We encode the textual and visual prompts by the pre-trained CLIP model~\cite{clip}, which allows us to train using a text prompt and perform inference using an image prompt in a zero-shot manner. Numerical prompts, \eg, 3D bounding boxes and locations, are projected by fully-connected layers into the same feature space as CLIP. With such unification, we do not distinguish different prompt formats anymore and this design enables the model to transfer knowledge between different prompts. The encoded task prompt is denoted as $\mathbf{t} \in \mathcal{R}^{T\times D}$, $T$ is the number of prompt tokens, and $D$ is the hidden dimension.


\subsection{3D Scene Encoding}
There are three widely used representations for 3D scenes: point clouds, voxel grids, and multi-view images, which have their unique advantages in different tasks. For example, voxels offer a uniform, grid-like structure ideal for instance segmentation~\cite{scannet200,mask3d}, point clouds provide spatial information crucial for visual grounding~\cite{3d-vista,vil3dref}, and multi-view images include a rich visual context beneficial for open-vocabulary learning~\cite{peng2023openscene,takmaz2023openmask3d}. 

To simultaneously handle various tasks in a 3D scene, we aim to achieve seamless unification of these representations. To accomplish this, we first group the dense 3D points into larger segments through unsupervised graph-based segmentation and masks~\cite{graph-seg}. 
The grouped segments are much fewer than the original points, making the training feasible by reducing the number of tokens in cross-attention. 
Then, we encode these scene representations by the corresponding encoders and pool the features to the segments in total of $M$.


\noindent \textbf{Point Cloud} To process the point cloud of a 3D scene, we first partition the full point cloud into the pre-generated segments. For each segment, we sample 1,024 points, normalize their coordinates into a unit sphere and then feed them into a pre-trained PointNet++ backbone~\cite{pointnet, pointnet++}  to obtain the point features, denoted by $\mathbf{P} = \{\mathbf{p}_0, \mathbf{p}_1,...,\mathbf{p}_M \} \in \mathcal{R}^{M \times D}$, where $M$ is the number of segments.


\noindent \textbf{Voxel}
To extract voxel features, we follow Mask3D~\cite{mask3d} to first discretize the 3D scene into voxels and then channel these voxels into a sparse convolutional U-net backbone. The sparse U-Net includes downsampling and upsampling stages to extract hierarchical information from the given scene. The extracted features of each voxel are remapped to the pre-generated segments and average-pooled to obtain segment-level features. We also apply a linear layer to project the voxel features to the hidden dimension $D$, denoted by $\mathbf{V} = \{\mathbf{v}_0, \mathbf{v}_1,...,\mathbf{v}_M \} \in \mathcal{R}^{M \times D}$.

\noindent \textbf{Multi-View Image}
Given multi-view images of a 3D scene, we follow OpenScene~\cite{peng2023openscene} to get the multi-view image features for each 3D point. We first compute per-pixel embeddings for each image using the pre-trained OpenSeg~\cite{openseg} segmentation model. We then back-project each 2D pixel into 3D point and aggregate the features from the associated multi-view pixels. Finally, we obtain the segment-level image features by simply average pooling, denoted as $\mathbf{I} = \{\mathbf{i}_0,\mathbf{i}_1,...,\mathbf{i}_M \} \in \mathcal{R}^{M \times D}$.

The final 3D scene representation is composed of these three segment-level features $\{\mathbf{V},\mathbf{I},\mathbf{P}\}$. Besides, we add positional encoding to segment-level features. In our implementation, We compute the average coordinate across all points within each segment as its 3D location, which is further encoded by an MLP into the same hidden dimension, denoted as $\mathbf{L} = \{\mathbf{l}_0,\mathbf{l}_1,...,\mathbf{l}_M \} \in \mathcal{R}^{M \times D}$. Unifying all features to the pre-generated segments is crucial for instance queries to effectively interact with these features in query learning.

\subsection{Prompt-guided Query Learning}
We propose a novel Transformer-like decoder to instruct the instance queries to assimilate scene and prompt information. This process begins with a set of instance queries $\mathbf{Q}_0$, whose values are initialized to zeros and positions are sampled via Farthest Point Sampling from 3D points~\cite{takmaz2023openmask3d}. Within the decoder layer $l$, the instance queries $\mathbf{Q}_{l}$ retrieve task-relevant information by first attending to the scene features $\{\mathbf{V},\mathbf{I},\mathbf{P}\}$ in parallel and then the task prompt $\mathbf{t}$, followed by a spatial self-attention as~\cite{3d-vista}. All attention layers are followed by the forward layer (FFN). The spatial self-attention utilizes location information from the coordinates of the farthest sampled points. Formally, we have:
\begin{align}
&\mathbf{Q}_{l}^{'} = \mathrm{FFN}(\mathrm{Norm}(\mathbf{Q}_l + \sum_{\mathbf{F} \in \{\mathbf{V},\mathbf{I},\mathbf{P}\}} \mathrm{MaskedCrossAttn}(\mathbf{Q}_l, \mathbf{F}))),  \\
&\mathbf{Q}_{l}^{''} = \mathrm{FFN}(\mathrm{Norm}( \mathbf{Q}_l^{'} + \mathrm{CrossAttn}(\mathbf{Q}_{l}^{'}, \mathbf{t}) )), \\
&\mathbf{Q}_{l+1} = \mathrm{FFN}(\mathrm{Norm}(\mathrm{SpatialSelfAttn}(\mathbf{Q}_l^{''}))).
\end{align}
Following~\cite{mask2former,mask3d}, we adopt masked attention when cross-attending to the scene features, which restricts the attention to localized features centered around the query for faster convergence and improved performance.  To support flexible inference when only some representations are available, we randomly drop out some scene features with rate 0.6 in masked-attention computation during training.  After $N$ decoder layers, we expect the final instance queries $\mathbf{Q}$ to have collected enough information for solving the given task.

\subsection{Output Heads and Losses}
We adopt the following three output heads to support a variety of 3D-VL tasks:
\noindent \textbf{Mask head} For each instance query, we apply a mask head to predict a binary mask over the pre-generated segments. Formally, we have: 
\begin{equation}
    p_\mathrm{mask} = \sigma(f_s (\mathbf{V} + \mathbf{I} + \mathbf{P}) \cdot f_q(\mathbf{Q})^T)
\end{equation}
where $f_s,f_q$ are projection layers. The dimension of $f_s(\mathbf{V} + \mathbf{I} + \mathbf{P})$ is $M \times D$ and $f_q(\mathbf{Q})$ is $Q\times D$, where $M, Q$, and $D$ represent the number of segments, queries, and hidden dimensions, respectively. The multiplication of $f_s$ and $f_q$ followed by sigmoid function $\sigma$ results in a binary mask in $M \times Q$ dimension. During training, we follow~\cite{mask3d,detr} to apply Hungarian Matching between queries and ground-truth objects, then calculate the mask loss:
\begin{equation}
\mathcal{L}_{\mathrm{mask}} = \lambda_{\mathrm{bce}} \mathcal{L}_{\mathrm{bce}} + \lambda_{\mathrm{dice}} \mathcal{L}_\mathrm{{dice}}
\end{equation}
where $\mathcal{L}_{\mathrm{bce}}$ is the binary cross-entropy loss and $\mathcal{L}_{\mathrm{dice}}$ is the Dice loss~\cite{takmaz2023openmask3d, mask2former, fastinst}.
    
\noindent \textbf{Grounding head} We apply a grounding head $f_g$ to predict if an instance query is related to the task. $f_g$ is implemented as linear projection layers. Formally, we have:
\begin{equation}
    p_\mathrm{grd} = \sigma(f_g(\mathbf{Q}))
\end{equation}
During training, if grounding labels are provided as supervision, we calculate a binary cross-entropy loss as the grounding loss $\mathcal{L}_{\mathrm{grd}}$.


\noindent \textbf{Generation head} We choose the decoder of a pre-trained T5-small~\cite{vlt5,t5} as the generation head to generate a text response, using all instance queries as the encoded inputs. During training, if text responses are provided as supervision for dense caption and QA task, we calculate the cross-entropy loss as the generation loss $\mathcal{L}_{\mathrm{gen}}$.


During training, the total loss is the weighted sum of losses from the above three heads:
\begin{equation} \label{eq:loss}
\mathcal{L}_{\text {total}}= \lambda_{\mathrm{mask}}\mathcal{L}_{\mathrm{mask}} + \lambda_{\mathrm{grd}} \mathcal{L}_{\mathrm{grd}} + \lambda_{\mathrm{gen}} \mathcal{L}_{\mathrm{gen}}
\end{equation}

\section{Experiments}

\subsection{Experimental setting}

\textbf{Training Datasets} \cref{tab:dataset} shows a summary of the datasets used for the multi-task training of \model. Notably, we combine eight datasets for training, including about 662K training samples for various tasks. 


\noindent \textbf{Training Details} The training procedure consists of two stages. In the first stage, we train the model with instance segmentation alone on ScanNet200 for 800 epochs. At this stage, instance segmentation is trained with a classification head on instance queries, instead of prompting object categories. At the second stage, we continue train the model on the full training set using the training objectives in \cref{eq:loss} for 50 epochs. We set the hidden dimension $D$ to 768, and query decoder layer $N$ to 4. We utilize the AdamW optimizer with a learning rate of 1e-4, batch size of 16, $\beta_1$ = 0.9, and $\beta_2$ = 0.98. The loss balance weights $\lambda_{\mathrm{mask}}, \lambda_{\mathrm{gen}}$ are set to 1, and $\lambda_{\mathrm{grd}}$ is set to 10. To further demonstrate the capability of \model, we also transfer it to an embodied agent for object navigation using the ObjNav task from CortexBench~\cite{majumdar2023we} and instruction-tune it with a large language model (LLM) Vicuna-7B using the instruction-following dataset from~\cite{huang2023embodied}. The whole training process is conducted on four NVIDIA A100 GPUs. More details can be found in the appendix.

\begin{table}[ht]
\caption{\textbf{Datasets for unified training.} The size of ScanNet200 is $\# \text{scenes (1202)} \times \#\text{categories (200)}$. }
\label{tab:dataset}
\centering
\small
\begin{tabular}{ccccr}
\toprule
Dataset & Task & Prompt & Heads & Size \\
\midrule
ScanNet200 \cite{scannet200} & instance segmentation & category & mask,grounding & 240K \\
ScanRefer \cite{scanrefer} & visual grounding & sentence & grounding & 37K \\
Nr3D \cite{referit3d} & visual grounding & sentence & grounding & 119K \\
Sr3D \cite{referit3d} & visual grounding & sentence & grounding & 66K \\
Multi3DRefer \cite{multi3drefer} & visual grounding & sentence & grounding & 44K \\
ScanQA \cite{scanqa} & question answering & question & grounding,generation & 30K \\
SQA3D \cite{sqa3d} & question answering & question & generation & 89K \\
Scan2Cap \cite{scan2cap} & dense captioning & 3D box & generation & 37K \\
\midrule
Total & - & - & - & 662K \\
\bottomrule
\end{tabular}
\end{table}


\subsection{Quantative Results}

\noindent \textbf{Instance segmentation on ScanNet200} As shown in \cref{tab:inst}, \model demonstrates SOTA performance for instance segmentation tasks on ScanNet200. Our approach achieves 20.2\% for $\mathrm{AP}$, 28.0\% for $\mathrm{AP}_{50}$, and 32.5\% for $\mathrm{AP}_{25}$ in promptable manner, which uses the output logits from the ground head after prompting all classes in ScanNet200. Notably, our methods surpass other open-vocabulary approaches, offering a more versatile language interface. However, our model's performance with tail classes is relatively less robust due to biases in the CLIP text encoder, which is analyzed in the appendix. To mitigate this issue, we incorporate a closed-vocabulary task head, which enables our model to surpass the closed-vocabulary SOTA method, Mask3D~\cite{mask3d}, especially in the tail classes.

\noindent \textbf{Zero-shot transfer to Replica}  In addition to evaluating our method's performance on ScanNet200, we investigate its generalization capabilities to an out-of-distribution dataset Replica~\cite{replica}. Our observations reveal that \model displays improved generalization abilities on unseen data, surpassing other open-vocabulary approaches in terms of $\mathrm{AP}$, $\mathrm{AP}_{50}$, and $\mathrm{AP}_{25}$ metrics by 0.2\%, 4.1\%, and 7.7\%. 
These findings imply a notable capability of \model for effective transfer to different datasets.

\begin{table*}[ht]
\caption{\textbf{Instance Segmentation results on the ScanNet200 validation set and zero-shot performance on Replica.}  The Average Precision (AP) is averaged over an overlapping range, and the $\mathrm{AP}_{50}$, $\mathrm{AP}_{25}$ is evaluated at 50\% and 25\% overlaps. Additionally, we provide AP scores for the head, common, and tail classes. The notation ``\model (\textit{w/cls})'' represents results using a closed-vocabulary classification head, whereas ``\model (\textit{prompt})'' denotes segmentation results in a promptable way, which corresponds to an open-vocabulary setting.}
\label{tab:inst}
\centering
\small
\begin{tabular}{l|cccccc|ccc}
\toprule 
\multirow{2}{*}{Model}  & \multicolumn{6}{c|}{ScanNet200} & \multicolumn{3}{c}{Replica} \\
 &  $\mathrm{AP}$ & $\mathrm{AP}_{50}$ & $\mathrm{AP}_{25}$ & head & common & tail & $\mathrm{AP}$ & $\mathrm{AP}_{50}$ & $\mathrm{AP}_{25}$ \\
\hline
\rowcolor[gray]{0.9} \textit{Closed-vocabulary} & & & & & & & & & \\
Mask3D ~\cite{mask3d}  & 26.9 & 36.2 & 41.4 & \textbf{39.8} & 21.7 & 17.9 & - & - & - \\
\model (\textit{w/cls})  & \textbf{27.0} & \textbf{38.9} & \textbf{46.3} & 35.8 & \textbf{24.2} & \textbf{20.0} & - & - & - \\
\hline
\rowcolor[gray]{0.9} \textit{Open-vocabulary} &  & & & & & & & & \\
OpenScene ~\cite{peng2023openscene}  & 11.7 & 15.2 & 17.8 & 13.4 & 11.6 & 9.9 & 10.9 & 15.6 & 17.3 \\
OpenMask3D ~\cite{takmaz2023openmask3d}  & 15.4 & 19.9 & 23.1 & 17.1 & 14.1 & \textbf{14.9} & 13.1 & 18.4 & 24.2 \\
\model (\textit{prompt})  & \textbf{20.2} & \textbf{28.0} & \textbf{32.5} & \textbf{30.9} & \textbf{17.0} & 11.3 & \textbf{13.3} & \textbf{22.5} & \textbf{31.9} \\
\bottomrule
\end{tabular}
\end{table*}

\noindent \textbf{Visual Grounding} \cref{tab:refer} provides an assessment of grounding accuracy for various methods on four benchmarks: ScanRefer, Nr3D, Sr3D, and Multi3DRefer. Our model consistently outperforms the other methods in most categories. On the ScanRefer, Nr3D, and Sr3D benchmarks, our model outperforms SOTA by 5.4\%, 2.3\%, and 3.3\%, respectively. 
Furthermore, on the Multi3DRefer benchmark, our model outperforms others in the ST (single target) and MT (multiple targets) categories and achieves the highest average score of 50.1\%. In the ZT (zero target) metric, our model lags behind the state-of-the-art. This performance gap could potentially be attributed to the fact that ZT is only present in the Multi3DRefer dataset. 
However, our model trained only on the Multi3DRefer dataset ``{\model} (\textit{sg.})'' exhibits better performance in the ZT and MT metric, but falls short of the unified trained model in other categories. 

\begin{table*}[ht]
\caption{\textbf{Grounding accuracy (\%) on 3D visual grounding benchmarks.} The results of ScanRefer and Multi3DRefer are reported under IoU@0.5. The results of Nr3D and Sr3D are reported using ground-truth masks during masked cross-attention. The ZT and ST results from Multi3DRefer are with distractors of the same class. ``\model (\textit{sg.})'' signifies a model trained on a single dataset.
}
\label{tab:refer}
\centering
\small
\resizebox{1.0\linewidth}{!}{
\begin{tabular}{l|ccc|ccc|ccc|cccc}
\toprule
\multirow{2}{*}{Method} & \multicolumn{3}{c|}{ScanRefer} & \multicolumn{3}{c|}{Nr3D} & \multicolumn{3}{c|}{Sr3D} & \multicolumn{4}{c}{Multi3DRefer} \\
               & Unique & Multiple & Avg. & Easy & Hard & Avg. & Easy & Hard & Avg. & ZT & ST & MT & Avg. \\
\midrule
ViL3DRel ~\cite{vil3dref}       & 68.6 & 30.7 & 37.7 & 70.2 & 57.4 & 64.4 & 74.9 & 67.9 & 72.8 & -    & -    & -    & -    \\
3DJCG ~\cite{3djcg}          & 64.3 & 30.8 & 37.3 & -    & -    & -    & -    & -    & -    & \textbf{66.9} & 16.7 & 26.2 & 26.6 \\
UniT3D ~\cite{unit3d}         & 73.1 & 31.1 & 39.1 & -    & -    & -    & -    & -    & -    & -    & -    & -    & -    \\
M3DRef-CLIP ~\cite{multi3drefer}    & 77.2 & 36.8 & 44.7 & 55.6 & 43.4 & 49.4 & -    & -    & -    & 39.4 & 30.6 & 37.9 & 38.4 \\
3D-VisTA ~\cite{3d-vista}      & 75.1 & 39.1 & 45.8 & 72.1 & 56.7 & 64.2 & 78.8 & 71.3 & 76.4 & -    & -    & -    & -    \\
\midrule
\model (\textit{sg.})  & 76.6 & 42.0 & 47.4 & 73.3 & 56.7 & 64.9 & 78.8 & 68.2 & 75.6 & 61.1 & 40.5 & \textbf{41.7} & 48.6 \\
\model         & \textbf{78.2} & \textbf{46.2} & \textbf{51.2} & \textbf{75.0} & \textbf{58.7} & \textbf{66.7} & \textbf{82.7} & \textbf{72.8} & \textbf{79.7} & 57.7 & \textbf{43.6} & 40.9 & \textbf{50.1} \\
\bottomrule
\end{tabular}
}
\end{table*}

\noindent \textbf{Question Answering}
On the ScanQA test set, \model outperforms all other methods in terms of the BLEU-1, METEOR, and CIDEr metrics. Specifically, our model surpasses SOTA by 8.6\% / 5.4\% for BLEU-1, 2.6\% / 1.0\% for METEOR, 11.2\% / 2.6\%  for CIDEr on ``w/ object'' and ``w/o object'' test set.  However, in terms of the EM@1 metric, the 3D-VisTA method outperforms our model with scores of 27.0\% ``w/ object'' and 23.0\% ``w/o object'', compared to our model's 26.1\% and 20.0\%, respectively. Different from 3D-VisTA, our model does not use a classification head for QA, which causes a performance drop in EM metric. 

\begin{table}[ht]
\centering
\small
\caption{\textbf{Answer accuracy on ScanQA.} Each entry denotes ``test w/ object" and ``test w/o object". EM@1 refers to the top 1 exact match accuracy, while BLEU-1, METEOR, and CIDEr denote text similarity scores between the predicted answer and the ground-truth answer. The notation "\model (\textit{sg.})" indicates a model trained on a single dataset rather than through unified joint training. }
\begin{tabular}{lcccc} 
\toprule
Method & EM@1 & BLEU-1 & METEOR & CIDEr \\
\midrule
ScanQA~\cite{scanqa}& 23.5 / 20.9 & 31.6 / 30.7 & 13.6 / 12.6 & 67.3 / 60.2 \\
3D-VisTA~\cite{3d-vista} & \textbf{27.0 / 23.0} & 34.4 / 30.2 & 15.2 / 12.9 & 76.6 / 62.6 \\
\midrule
\model (\textit{sg.})       & 18.9 / 16.1 & 34.7 / 30.5 & 14.5 / 12.1 & 69.3 / 56.0 \\
\model              & 26.1 / 20.0 & \textbf{43.0 / 36.1} & \textbf{17.8 / 13.9} & \textbf{87.8 / 65.2} \\
\bottomrule
\end{tabular}
\end{table}

For the SQA3D task, it is worth noting that our proposed model falls slightly behind the SOTA in SQA3D, with a difference of 1.4\%. As our model utilizes the CLIP text encoder, it may face limitations in understanding long sentences. 

\begin{table}[ht]
\centering
\small
\caption{\textbf{Answer accuracy on SQA3D under  question types.} 
}\label{tab:sqa3d}
\begin{tabular}{lccccccc}
\toprule
\multirow{2}{*}{Method} & \multicolumn{6}{c}{Test set} & \multirow{2}{*}{Avg.} \\ 
\cline{2-7}
& What \quad & Is \quad & How \quad  & Can \quad  & Which \quad  & Other \quad  & \\ 
\hline
ClipBERT \cite{sqa3d} & 30.2 & 60.1 & 38.7 & 63.3 & 42.5 & 42.7 & 43.3 \\
SQA3D \cite{sqa3d} & 31.6 & \textbf{63.8} & \textbf{46.0} & 69.5 & 43.9 & 45.3 & 46.6 \\
3D-VisTA \cite{3d-vista} & 34.8 & 63.3 & 45.4 & \textbf{69.8} & \textbf{47.2} & \textbf{48.1} & \textbf{48.5} \\  
\midrule 
\model (\textit{sg.}) & 35.6 & 62.7 & 45.2 & 66.3 & 43.3 & 43.3 & 46.8 \\  
\model & \textbf{37.1} & 61.3 & 44.5 & 60.9 & 47.0 & 45.1 & 47.1 \\  
\bottomrule
\end{tabular}
\end{table}

 \noindent \textbf{Dense Captioning} For the dense captioning task, \model outperforms all other models in the CIDEr, METEOR, and ROUGE metrics. With a CIDEr score of 80.3\%, it significantly surpasses the next best, 3D-VisTA, which has a CIDEr score of 66.9\%. Importantly, the performance of \model trained on multiple tasks and datasets exceeds that of \model trained on a single task and dataset, showcasing the effectiveness of multi-task joint training.
 

\noindent \textbf{Object Navigation}
To further verify \model's effectiveness, we finetune it on the ObjNav task from CortexBench~\cite{majumdar2023we} for navigating an embodied agent to a desired object. The proposed \model provides global 3D features to the navigation agent that can improve the baseline VC-1 by a significant margin, achieving a 22.9\% increase in success rate. Importantly, our method showcases consistent and stable performance, even in the absence of a compass or directional guidance.

\begin{table}[htbp]
\begin{minipage}[t]{0.5\linewidth} 
\centering
\caption{\textbf{Captioning results on Scan2Cap} under IoU@0.5  with various text similarity scores. "\model (\textit{sg.})" indicates a model trained on a single dataset rather than through unified training.
}
\label{tab:scan2cap}
\resizebox{\linewidth}{!}{
\begin{tabular}{lcccc}
\toprule
Method & CIDEr \quad & BLEU-4 \quad & METEOR \quad  & ROUGE \quad  \\ \midrule
Scan2Cap \cite{scan2cap} & 35.2 & 22.4 & 21.4 & 43.5 \\
3DJCG \cite{3djcg}   & 47.7 & 31.5 & 24.3  & 51.8 \\
3D-VisTA ~\cite{3d-vista} & 66.9 & 34.0 & 27.1 & 54.3 \\ 
\midrule
\model (\textit{sg.}) & 75.6 & 34.4 & 28.6 & 57.1 \\
\model & \textbf{80.3} & \textbf{36.0} & \textbf{29.1} & \textbf{57.9} \\
\bottomrule
\end{tabular}
}
\end{minipage}
\hfill
\begin{minipage}[t]{0.46\linewidth} 
\caption{\textbf{Results on ObjNav from CortexBench}~\cite{majumdar2023we}. Note we reproduce the result of  ``VC-1 (ViT-B)'' ourselves due to the slight mismatch we have found. Only variants with {\model} use 3D input. The result with $^*$ is also without a compass sensor.
}
\label{tab:objnav}
\resizebox{1\linewidth}{!}{
\begin{tabular}{lccc}
\toprule
Model & Success $\uparrow$ \quad &  SPL $\uparrow$ \quad & Soft-SPL $\uparrow$ \quad  \\ \midrule
VC-1 (ViT-B)~\cite{majumdar2023we} & 57.1  &  0.31 & 0.41  \\
\model & \textbf{80.0} & \textbf{0.50} & \textbf{0.60} \\
\model w/o GPS$^*$ & 75.0 & 0.45 & 0.50  \\ \bottomrule
\end{tabular}
}
\end{minipage}
\end{table}


\subsection{Ablation study}
\noindent \textbf{Query Decoder Depth}
In this study, we examine the influence of decoder depth on downstream tasks. The results suggest that a 4-layer decoder outperforms both 2-layer and 6-layer ones on all tasks. Consequently, we choose a 4-layer query decoder for \model.

\begin{figure}[th]
    \centering \label{fig:decoder}
    \includegraphics[width=0.95\linewidth]{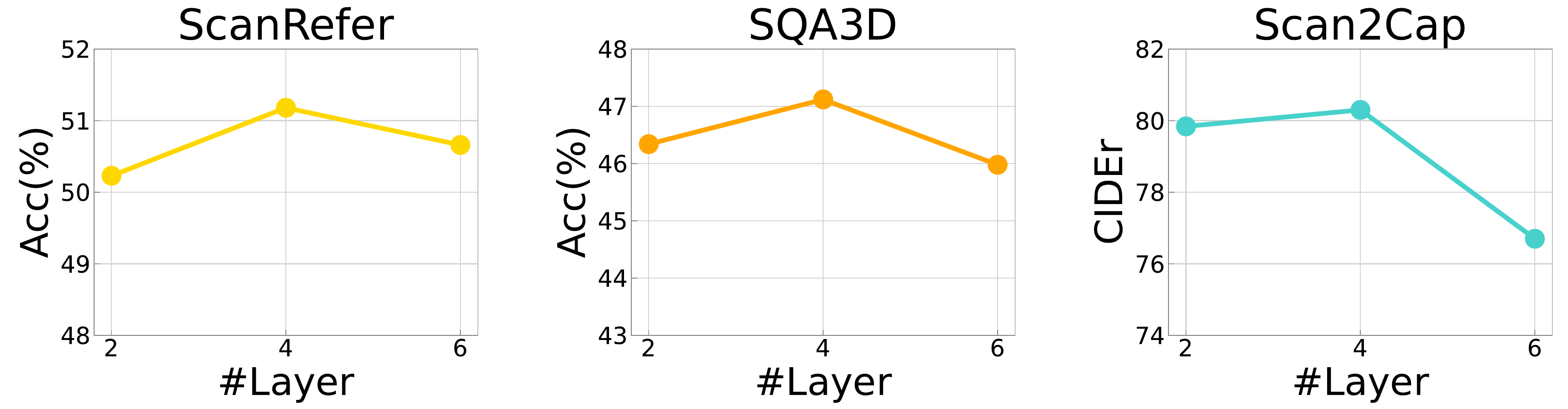}
    \caption{\textbf{Ablation study of query decoder depth.}}
\end{figure}

\noindent \textbf{Scene Features}
Our result in \cref{tab:ablation_feat} reveals the benefits of incorporating voxel, point, and image features on refering (ScanRefer Acc), question answering (SQA3D Acc), and captioning (Scan2Cap CIDEr) tasks. Specifically, the addition of point features results in performance changes of +3.1\%,  +1.7\%, +6.8\% and while the inclusion of image features, based on voxel and point, leads to improvements of +2.0\%, +1.7\%, and +5.7\% on these respective tasks. These results exemplify the effectiveness of our feature alignment approach, as it enables enhanced performance across grounding, QA, and captioning tasks. 

\noindent \textbf{Flexible Inference} 
To assess how our model performs under the constraints of limited scene representations, we conduct inference while omitting certain scene features, using a single set of model weights for testing. From \cref{tab:ablation_feat}, we can observe that \model achieves comparable performance with the model trained with specific scene features when the image feature is excluded. When both image and point features are absent, the \model outperforms the specific-tuned model, demonstrating the improved generalization ability through training with multiple representations.


\noindent \textbf{Influence between tasks. } 
In  \cref{tab:ablation_task}, we examine the interplay among Refer, QA, and captioning tasks. Our findings indicate that incorporating data from the Refer task yields improvements of +1.8\% for QA and +2.2\% for captioning. While adding QA data shows no significant benefit for Refer, it does contribute to a +1.2\% enhancement in captioning performance. Conversely, data from the captioning task positively impacts both Refer and QA tasks, with gains of +0.6\% and +0.7\%, respectively.

\begin{table}[htbp]
\centering
\begin{minipage}[t]{0.6\linewidth} 
\centering
\caption{\textbf{Ablation study of scene features}. Each entry denotes \model ``trained with specific scene features'' and ``trained with all features but some removed during inference''. These results indicate that \model can support flexible inference.}  \label{tab:ablation_feat}
\resizebox{\linewidth}{!}{
\begin{tabular}{cccccc}
\toprule
Voxel  & Point  & Image  & Refer  & QA  & Caption   \\ \hline
\checkmark & & & 46.1 / 47.1 & 43.7 / 44.2 & 67.8 / 68.1 \\
\checkmark & \checkmark & & 49.2 / 49.4 & 45.4 / 45.8 & 74.6 / 74.7 \\
\checkmark & \checkmark & \checkmark & 51.2  & 47.1  & 80.3  \\
\bottomrule
\end{tabular}
}
\end{minipage}
\hfill
\begin{minipage}[t]{0.38\linewidth} 
\centering
\caption{\textbf{Ablation study of influence between tasks.} Each entry denotes performance gain by introducing extra task data for joint training.}  \label{tab:ablation_task}
\resizebox{\linewidth}{!}{
\begin{tabular}{lccc}
\toprule
Task Data  & Refer  & QA  & Caption \\ \hline
+Refer & -  & \inc{1.8}   & \inc{2.2}  \\
+QA & \inc{0.0}  & -   & \inc{1.2}  \\
+Caption & \inc{0.6}  & \inc{0.7}   & -  \\
\bottomrule
\end{tabular}
}
\end{minipage}
\end{table}



\subsection{Qualitative results}
\begin{figure*}[ht]
    \centering
    \includegraphics[width=1\linewidth]{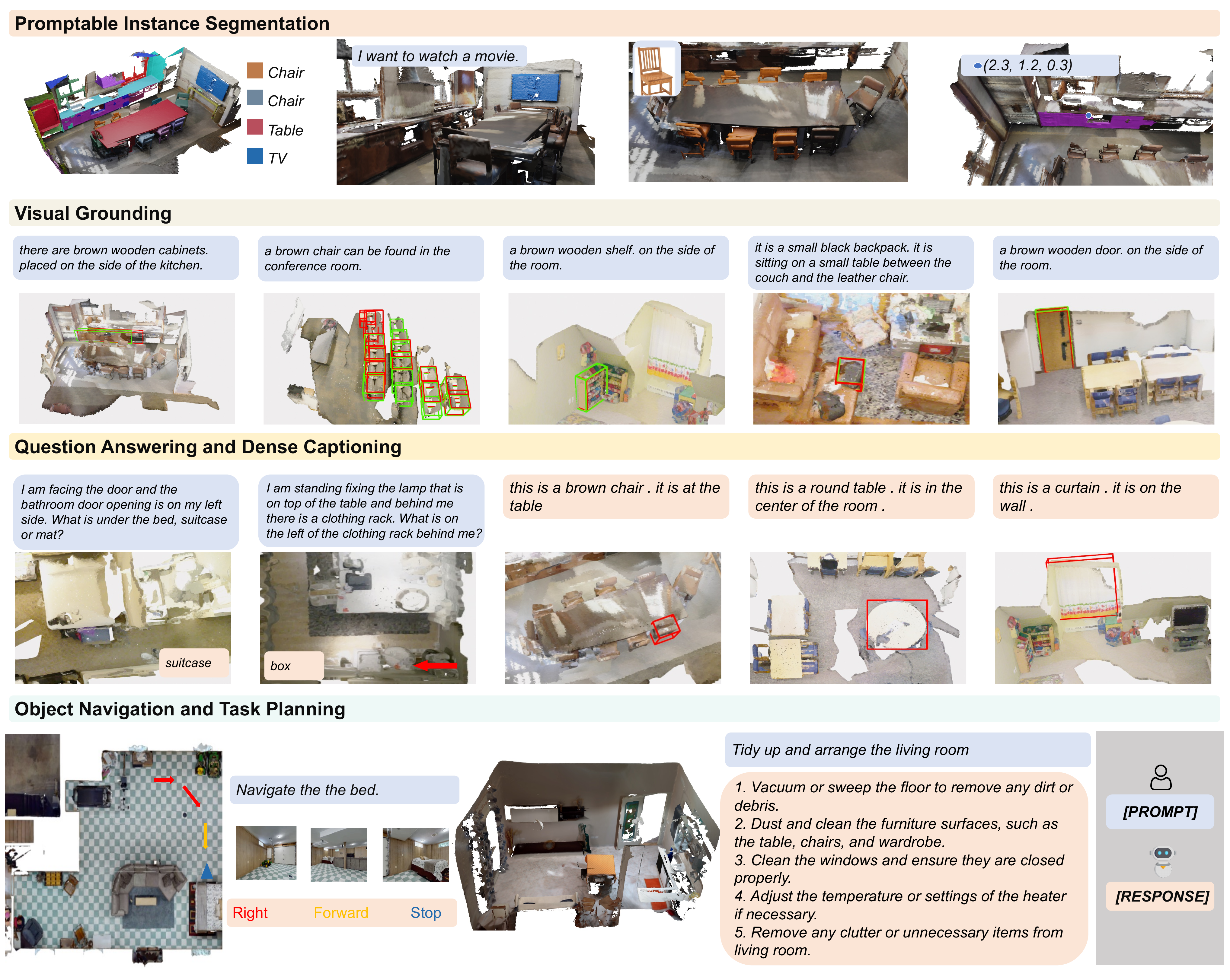}
    \caption{\textbf{Qualitative examples from \model. }Red bounding box denotes the result from \model, and green denotes ground truth.}
    \label{fig:qualitative}
\end{figure*}

\cref{fig:qualitative} presents the qualitative results of our model, \model. The first row displays the outcome of promptable segmentation, which includes prompt forms like class names, a sentence, an image, and a 3D location. The second and third rows demonstrate the results of 3D-VL tasks, encompassing visual grounding, question answering, and dense captioning. These results exhibit the model's capabilities in solving various 3D-VL tasks. The last row illustrates the model's proficiency in object navigation and task planning. \model successfully navigates towards the intended destination and formulates a plan to organize a living room. These qualitative results emphasize that \model is a unified model with the potential to be applied to more embodied agent tasks as a fundamental 3D encoding module.

\section{Conclusions and Future Works}
In conclusion, our proposed \model addresses the challenges in 3D vision-language learning (3D-VL) by offering a unified approach that integrates multiple representations and supports a wide range of tasks. By leveraging the cross-attention mechanism between instance queries and representations, our model generates task-aware instance queries, guided by prompts specific to each task. Through extensive experiments on various 3D-VL benchmarks, we demonstrate the effectiveness of unifying different representations in achieving state-of-the-art results. Notably, our model demonstrates promotable 3D instance segmentation, which contributes to advancing open-world 3D scene understanding. With these accomplishments, \model may potentially impact embodied intelligence more, representing a step towards aligning natural language with the 3D physical world.

At present, the scale and generalization capability of \model still exhibits a gap compared to 2D vision language foundational models. We aim to enhance \model by scaling it with more scenes and language data. Additionally, we plan to utilize the 2D foundational model to guide 3D training.

\noindent \textbf{Acknowledgements}
We thank the anonymous reviewers for their constructive suggestions. Their insights have greatly improved the quality and clarity of our work. This work was partially supported by the National Science and Technology Major Project (2022ZD0114900). Additionally, Ziyu Zhu and Zhidong Deng were supported by the National Science Foundation of China (NSFC) under Grant No. 62176134 and by research and application on AI technologies for smart mobility funded by SAIC Motor.

\bibliographystyle{splncs04}
\bibliography{main}

\newpage



\newpage
\appendix
\renewcommand\thefigure{A\arabic{figure}}
\setcounter{figure}{0}
\renewcommand\thetable{A\arabic{table}}
\setcounter{table}{0}
\renewcommand\theequation{A\arabic{equation}}
\setcounter{equation}{0}
\pagenumbering{Alpha}
\renewcommand*{\thepage}{A\arabic{page}}
\setcounter{footnote}{0}

{\Large{\noindent\textbf{Appendix}}}

\section{Implementation Details}
This section introduces more details about model architecture, training details, and downstream task settings.
\subsection{Model architecture}

\noindent \textbf{Prompt Encoding} To encode textual prompts, we utilize the CLIP (clip-vit-large-patch14) text encoder with a projection layer. This allows us to encode text and images into a joint embedding space. For numerical prompts, we employ a linear layer to project locations and bounding box sizes into $\mathbb{R}^D$ space. This projection aligns the feature space with CLIP. When it comes to image prompts, our goal is to utilize image semantics to replace text prompts for zero-shot transfer. To achieve this, we encode the text in the form of ``\texttt{[SOS]} \texttt{[object]}  \texttt{[EOS]}'', where \texttt{[object]} is substituted with image semantic features encoded by the CLIP image encoder. This approach allows us to leverage the semantic information contained within images and use it as a replacement for traditional text prompts, facilitating zero-shot transfer learning.

\noindent \textbf{Scene Representation}
For point cloud encoding, we utilize a three-layer PointNet++~\cite{pointnet++} architecture. This architecture operates on the input points with different radii: 0.2, 0.4, and sampling all points within this ball. The output of this PointNet++ is an aggregated 768-dimensional feature representation. For voxel encoding, we use MinkowskiRes16UNet34C~\cite{minkowski} with a voxel size of 0.02m. It combines the ResNet-16 backbone with the UNet architecture in the Minkowski Engine framework. It applies 3D convolutions and pooling operations to extract hierarchical features from the input voxelized data. The network progressively downsamples the spatial resolution in the encoder path and upsamples it in the decoder path, while capturing and fusing features at multiple scales through skip connections. We use four levels of voxel features and project them all to dimension 768. For image encoding, we use a pre-trained OpenSeg feature backbone~\cite{openseg}. 

\noindent \textbf{Prompt-guided Query Learning}
In query initialization, we first use furthest point sampling~\cite{mask3d} to sample points from original point clouds. After we acquire the location of sampled points, we apply Fourier spatial position encoding~\cite{fourier-pos} to encode points $\mathbf{p}_i$ as follows:
\begin{equation}
 \mathbf{l_i}=\left[\cos (2\pi \mathbf{p}_i W_f^T) \| \sin (2\pi \mathbf{p}_i W_f^T)\right]
\end{equation}
During masked cross-attention, the location information $\mathbf{l}$ is incorporated into the computation of attention by adding it to the queries $\mathbf{\tilde{Q}}_{l}$ and keys $\mathbf{\tilde{F}}$. Here, $\mathbf{\tilde{Q}}_{l}$ represents the instance queries, and $\mathbf{\tilde{F}}$ denotes the features augmented by locations. Following the approach described in \cite{mask3d}, we enforce attention restriction using a mask generated from the previous layer, denoted as $\mathbf{M}^{l-1}$. This mask ensures that instance queries can only attend to voxel, image, and point features within their corresponding intermediate instance mask predicted by the previous layer. This masked-attention mechanism is constrained to focus on relevant features within the instance mask, which helps in capturing context and information specific to each instance, preventing cross-contamination between different instances during the attention computation.

\begin{equation}
\mathbf{Q}^{'}_{l}=\operatorname{softmax}\left(W^T_{q}\mathbf{\tilde{Q}}_{l} \mathbf{\tilde{F}}^T W^T_{k} / \sqrt{D}+\mathbf{M}_{l-1}\right) W_v \mathbf{F}
\end{equation}

\noindent \textbf{Output Head} The mask head employs two separate MLPs for queries and features, using a hidden dimension of 768 to map them to a shared space~\cite{mask3d}. The grounding head consists of a two-layer MLP with a hidden dimension of 384 and produces a probability score as output~\cite{3d-vista}. The generation head uses a pre-trained T5-small~\cite{t5}.

\subsection{Training settings} In this section, we provide a detailed explanation of our training procedure. Image features are preserved before training, and only voxel and point features are trained. For the unsupervised generation of segment masks, we maintain the same hyperparameters as those used in ScanNet, which are applied to both ScanNet and Replica datasets. For data augmentation, we utilize techniques such as horizontal flipping, random rotations around the z-axis, elastic distortion, and random scaling. We also implement color augmentations including jittering, as well as brightness and contrast adjustments.  We set the number of queries to 120 to strike a balance between speed and the number of objects present in the scene.  During the training phase of instance segmentation, we adopt the use of ground truth mask guidance  ~\cite{fastinst} for the initial 200 epochs to expedite the model's convergence.  We employ a cosine decay model for our learning rate scheduler. During instruction tuning, we replace the T5 decoder with Vicuna and conduct 10 epochs of fine-tuning. We use a learning rate of 3e-5 and train the model using 4 A100 GPUs. For object navigation, we connect our query decoder to a recurrent neural network and use the AdamW optimizer with a learning rate of 1e-3. The object navigation agent is trained in the HM3D environment for 100 million steps with 16 parallel environments.

\subsection{Details for Each Task}
\noindent \textbf{Instance Segmentation}
In our instance segmentation experiments, we utilize the ScanNet200 dataset, which is an extension of the ScanNet dataset. ScanNet is an annotated dataset that comprises 3D reconstructed indoor scenes, offering a rich representation of various room types such as hotels, libraries, and offices. The dataset is divided into three distinct subsets: 1,202 scenes for training, 312 scenes for validation, and 100 scenes for the hidden test set. In our evaluation, we utilize the mean average precision (mAP) metric at various Intersections over Union (IoU) thresholds. This metric provides a comprehensive assessment of instance segmentation performance. In closed vocabulary experiments, we employ 120 queries and select the top 200 output masks based on their classification scores.  In open vocabulary experiments, we prompt all 200 classes and select the top 200 masks using binary cross-entropy (BCE) logits. This setup allows for the exploration of a broader range of object classes, extending beyond the limitations of a closed vocabulary.

\noindent \textbf{Visual Grounding}
In our experiments, we evaluate our model on three different datasets for visual grounding tasks: ScanRefer, Nr3D/Sr3D, and Multi3DRefer.

ScanRefer dataset consists of 51,583 sentences written by humans to describe 800 scenes from the ScanNet dataset. The dataset is categorized into unique and multiple subsets based on whether the target object described in the sentence is a unique class within the scene. The evaluation metric for this task is accuracy, measured under intersection over union (IoU) thresholds of 0.25 and 0.5. In this task, we aim to predict the location of the target object described in the sentence.

Nr3D/Sr3D datasets comprise synthetic and human utterances, respectively. Sr3D contains 83,572 synthetic utterances, while Nr3D contains 45,503 human utterances. Both datasets are split into ``Easy''/``Hard'' and ``ViewDep''/``ViewIndep'' subsets. The ``Hard'' samples involve scenes with more than two distractors. The evaluation metric for these datasets is based on the F1 score, measured at IoU thresholds of 0.25 and 0.5. Ground truth masks and locations are used for evaluation, following the original settings of these benchmarks to ensure fair comparison.

Multi3DRefer dataset consists of 61,926 language samples, categorized into zero-target (ZT), single-target (ST), and multi-target (MT) referring samples. In this dataset, we predict masks for the referred objects. The evaluation metric used is the F1 score at IoU thresholds of 0.25 and 0.5. Again, we follow the original settings of this benchmark to ensure fair and consistent evaluation.

\noindent \textbf{Question Answering}
ScanQA is a 3D question answering dataset with 41,363 questions and 58,191 answers, focusing on spatial relations. Evaluation metrics include exact matches (EM@1, EM@10) and text similarity metrics (BLEU-1, ROUGE, METEOR, CIDEr). We also evaluate \model on SQA3D, a benchmark for scene understanding with 6.8k situations and 33.4k diverse questions. The evaluation metric in SQA3D is answer accuracy across different question types. Both benchmarks use a generation approach for answering questions.

\noindent \textbf{Dense Captioning}
Scan2Cap is a dense captioning benchmark that utilizes texts from the ScanRefer dataset. In this benchmark, we evaluate the performance of models in generating dense captions for 3D scenes. To assess the quality of the generated captions, we report text similarity scores, including CIDEr, BLEU-4, METEOR, and ROUGE, under different IoU (Intersection over Union) scores. These metrics provide a measure of the similarity between the generated captions and the reference captions, taking into account the level of overlap between the predicted and ground truth regions of interest.

\noindent \textbf{Object Navigation}
We utilize the object navigation benchmark from CortexBench, which is based on the HM3D-SEM dataset. The dataset consists of 80 training scenes, 20 validation scenes, and 20 test scenes from the HABITAT platform. For our evaluation, we use the validation split for comparison purposes. In this benchmark, the agent is modeled after the LocoBot, and the sensors are positioned at the top of the agent's head. The RGB camera used has a resolution of 640×480 and a horizontal field of view of 79 degrees. The objective for the agent is to navigate within the environment and locate objects belonging to one of six categories: ``chair'', ``bed'', ``plant'', ``toilet'', ``tv/monitor'', and ``sofa''. The agent has a maximum of 500 steps to complete the task successfully. The determination of successful episodes is based on the agent stopping within 0.1m of a viewpoint that is within 1m of any instance of the target object. This proximity criterion is used to determine if the agent has successfully located the target object.

\noindent \textbf{Task Planning}
To demonstrate the task planning ability of our model, we employ the LEO~\cite{huang2023embodied} instruction following data for fine-tuning. This dataset comprises scenes from 3RScan and ScanNet, enriched with 3D dialogues, scene-aware task planning information, and paired 3D question answering data. By utilizing this data, we aim to enhance our model's capability to understand instructions, follow them in a 3D environment, and effectively plan and execute tasks based on the given instructions. The inclusion of scene-aware task planning information allows our model to incorporate contextual knowledge about the environment and optimize its decision-making process accordingly. Furthermore, the paired 3D question answering data enables our model to not only follow instructions but also answer questions related to the scene and the performed tasks.

\section{Extra Ablation Studies}
\begin{figure}[!h]
    \centering
    \includegraphics[width=0.6\linewidth]{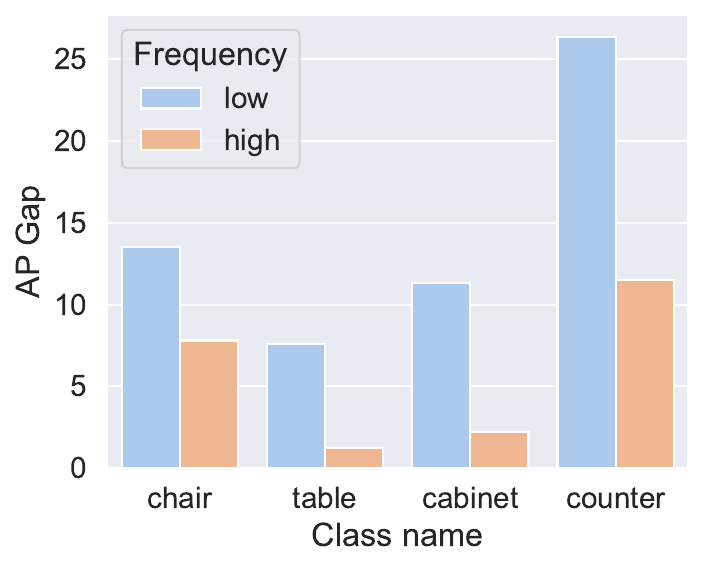}
    \caption{Differences in classes with similar semantics but have different frequencies in the ScanNet200 dataset. AP Gap denotes the difference between closed vocabulary and open vocabulary.}
    \label{fig:bar-class-dif}
\end{figure}

\noindent \textbf{Bias in CLIP text encoder}
From \cref{fig:bar-class-dif}, we find that when comparing classes with similar semantics, such as ``cabinet'' and ``kitchen cabinet'', or ``chair'' and ``armchair'', we observe differences in the instance segmentation quality between a closed vocabulary head and an open vocabulary approach. In the closed vocabulary setting, where specific class labels are predefined, the instance segmentation quality tends to be higher. This is because the model is trained to recognize and segment instances based on the pre-defined class labels, leading to more accurate and precise results. On the other hand, in the open vocabulary setting, where the model has the flexibility to generate class labels on the fly, the instance segmentation quality may vary. The model may struggle to distinguish between closely related classes, resulting in less accurate segmentation boundaries and potentially merging instances that should be separated. 

\begin{figure}[!h]
    \centering
    \includegraphics[width=0.6\linewidth]{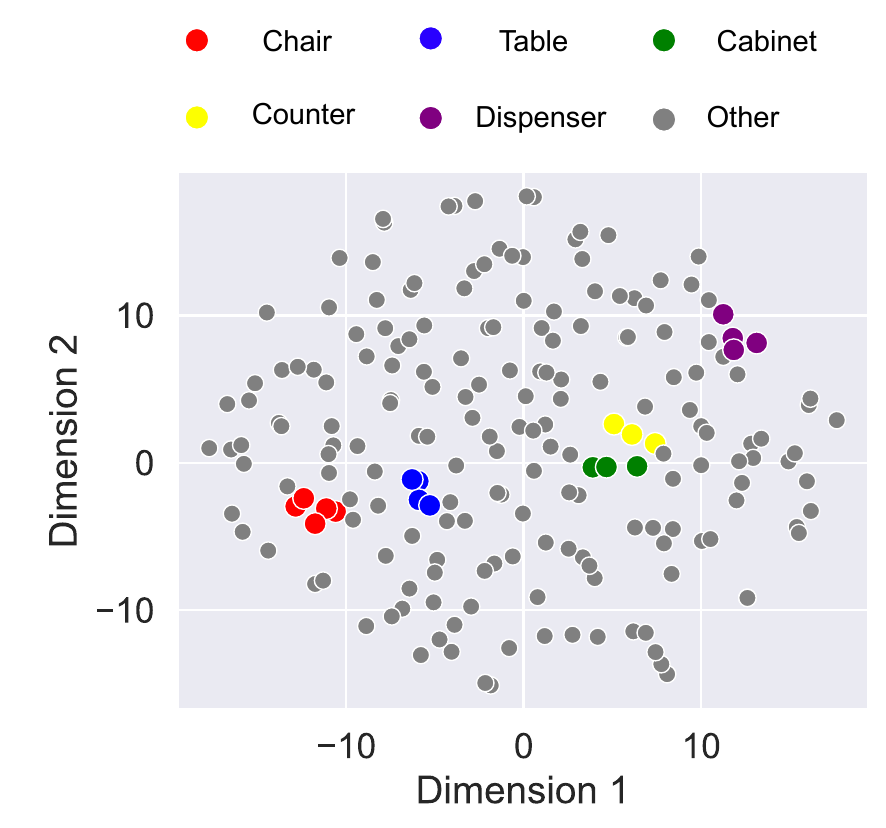}
    \caption{t-SNE visualization of ScanNet200 class names encoded by CLIP text encoder. }
    \label{fig:tsne}
\end{figure}

These differences arise from the challenge of disambiguating between classes that have similar meanings but subtle text differences in CLIP embedding space as shown in \cref{fig:tsne}. The closed vocabulary approach benefits from pre-defined class boundaries, while the open vocabulary approach relies on the model's ability to generalize and make fine-grained distinctions between similar classes. Therefore, when dealing with classes that have similar semantics but differ in instance segmentation quality, the choice between closed vocabulary and open vocabulary approaches should be carefully considered, taking into account the specific requirements and trade-offs of the task at hand.

\noindent \textbf{Ablation of structures}
Apart from our proposed decoder structure in \cref{fig:model}, we explore two alternative structures for the cross-attention process: The parallel structure and the sequential structure. In the parallel structure, instance queries cross-attend to all visual features and the encoded task prompt in parallel:
\begin{align}
&\mathbf{Q}_{l}^{'} = \sum_{\mathbf{F} \in \{\mathbf{V},\mathbf{I},\mathbf{P},\mathbf{t}\}} \mathrm{MaskedCrossAttn}(\mathbf{Q}_l, \mathbf{F}) \\ &\mathbf{Q}_{l+1} = \mathrm{SpatialSelfAttn}(\mathbf{Q}_l^{'})
\end{align}
Conversely, in the sequential structure, instance queries cross-attend to all visual features and the encoded task prompt in sequence:
\begin{align}
&\mathbf{Q}_{l}^{'} = \mathrm{MaskedCrossAttn}(\mathbf{Q}_l, \mathbf{V}) \\ &\mathbf{Q}_{l}^{''} = \mathrm{MaskedCrossAttn}(\mathbf{Q}_l^{'}, \mathbf{I}) \\ &\mathbf{Q}_{l}^{'''} = \mathrm{MaskedCrossAttn}(\mathbf{Q}_l^{''}, \mathbf{P}) \\ 
&\mathbf{Q}_{l}^{''''} = \mathrm{CrossAttn}(\mathbf{Q}_l^{'''}, \mathbf{t}) \\ 
&\mathbf{Q}_{l+1} = \mathrm{SpatialSelfAttn}(\mathbf{Q}_l^{''''})
\end{align}
From \crefrange{tab:ScanRefer}{tab:Scan2Cap},  it is evident that the structure outlined in our main paper delivers superior performance compared to both parallel and sequential approaches.

\section{More qualitative results}
\begin{figure}[!h]
    \centering
    \includegraphics[width=0.95\linewidth]{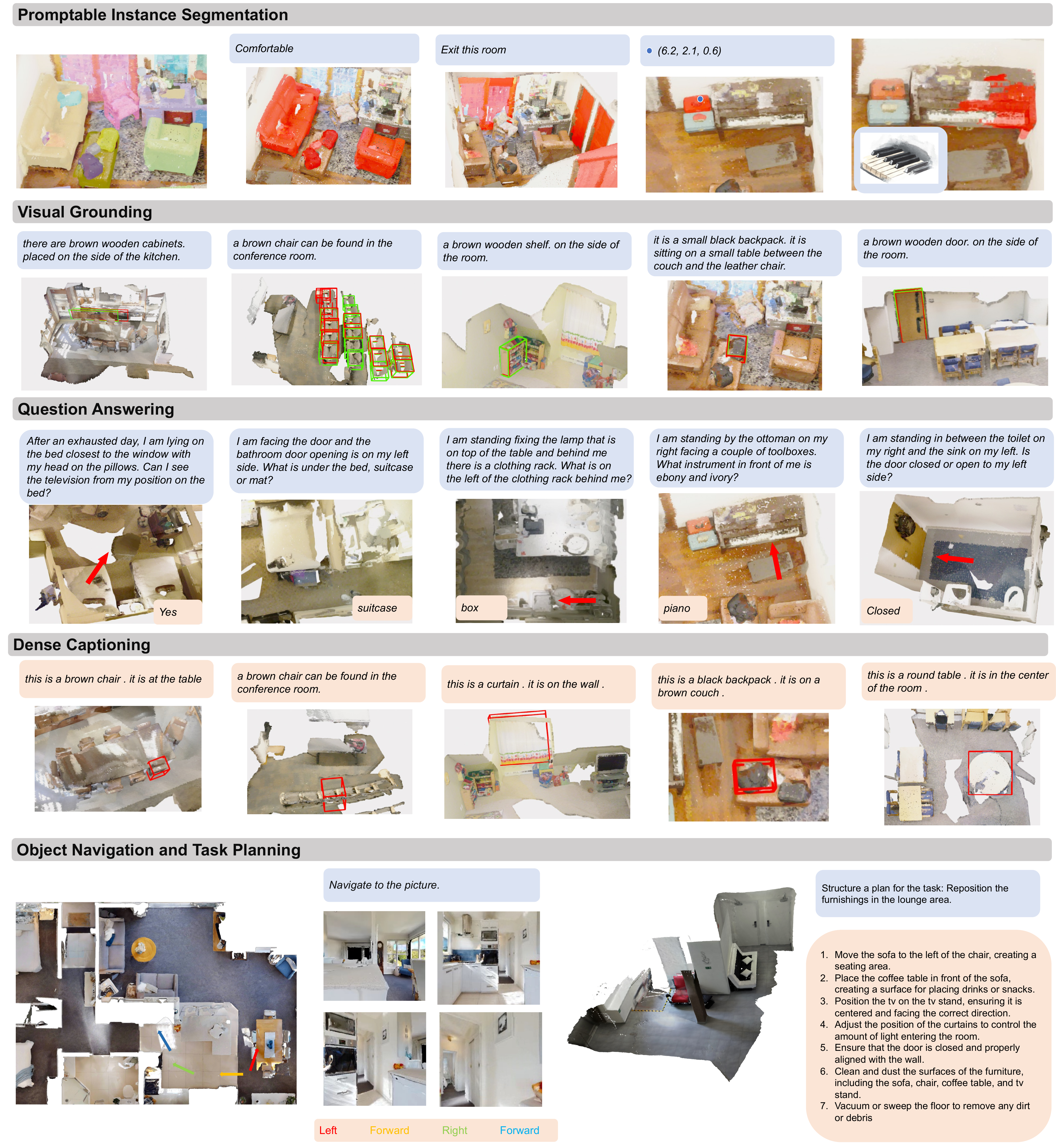}
    \caption{More visualization on promptable segmentation, visual grounding, question answering, dense captioning, object navigation, and task planning. \textcolor{red}{Red} bounding box denotes predicted result, \textcolor{green}{green} denotes groundtruth result.}
    \label{fig:more-vis}
\end{figure}

\noindent \textbf{More visualization}
In \cref{fig:more-vis}, we present results from various tasks conducted on a range of scenes. In the case of promptable instance segmentation, \model is capable of producing high-fidelity masks and reasoning about affordances such as comfort and exit. Interestingly, when an image prompt featuring a part of a piano keyboard is provided, \model segments not the entire piano but a small section of the keyboard. This suggests that the query can extract semantic information from the image prompt and segment a specific part of an instance. In the visual grounding task, it is observed that \model is capable of grounding both single and multiple objects, and it demonstrates an understanding of the relations between objects. In the question answering task, \model exhibits an understanding of the surrounding environment and can infer object classes and states based on the given situation. In the dense captioning task, our model can describe an object and its relationships with surrounding objects. In object navigation and planning, the examples presented demonstrate our model's capacity to support embodied agent tasks.

\begin{figure}[!h]
    \centering
    \includegraphics[width=0.95\linewidth]{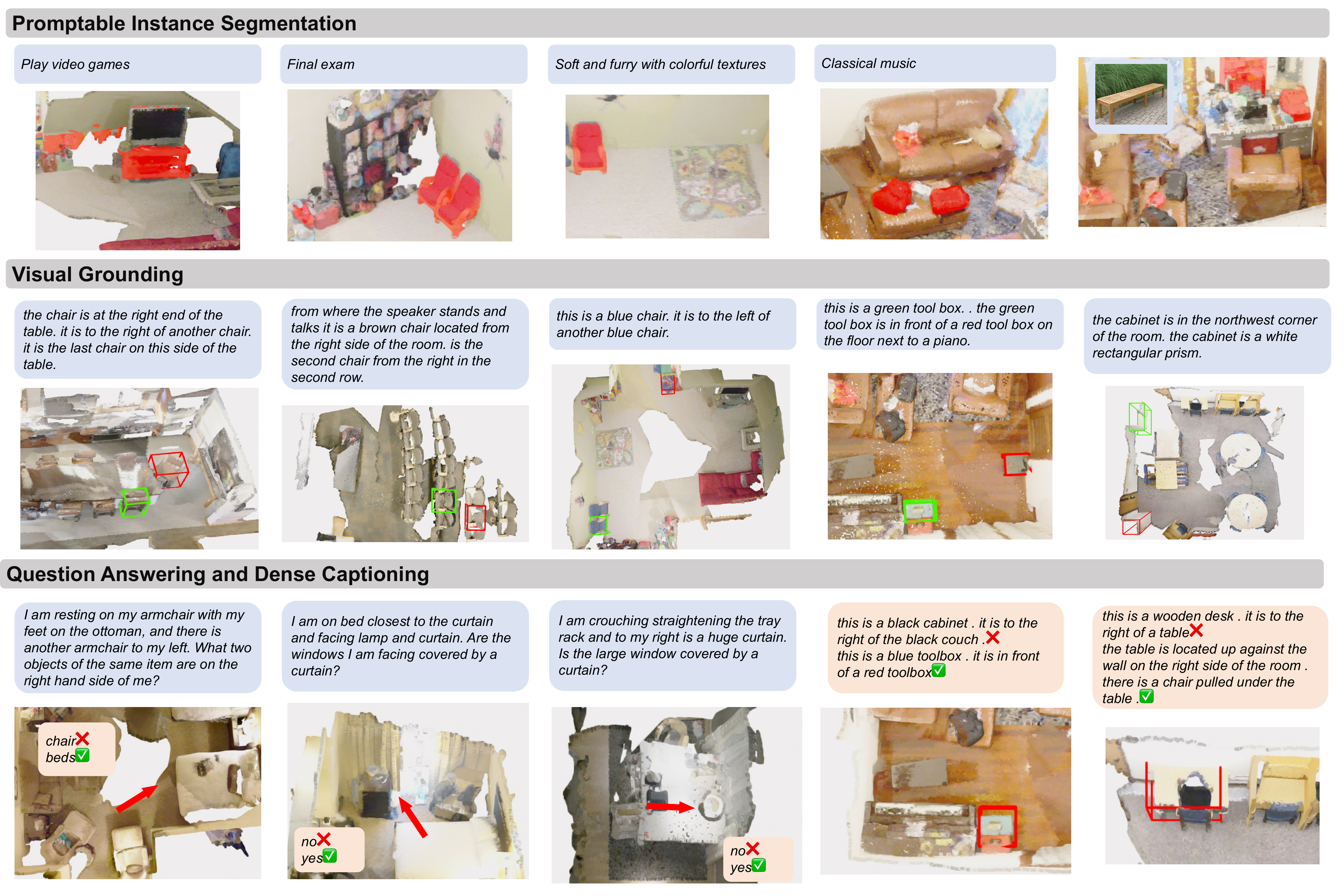}
    \caption{Failure cases in promptable segmentation, visual grounding, question answering, and dense captioning.}
    \label{fig:failure-case}
\end{figure}

\noindent \textbf{Failure cases}
We present failure cases of \model in \cref{fig:failure-case}. It can be observed that \model may segment unrelated objects when text prompts lack sufficient clarity, such as in the case of prompts like ``classical'', ``exam'', or ``video games''. Additionally, when image prompts contain background noise, it may lead to confusion for the model, causing it to segment random objects across different regions in a scene. In visual grounding and question answering tasks, the model may struggle to recognize complex spatial relations and comprehend long sentences. This limitation can potentially be mitigated by employing a more powerful text encoder. In dense captioning tasks, the model may face challenges in accurately understanding small objects in terms of their semantics.

\begin{figure}[!h]
    \centering
    \includegraphics[width=0.95\linewidth]{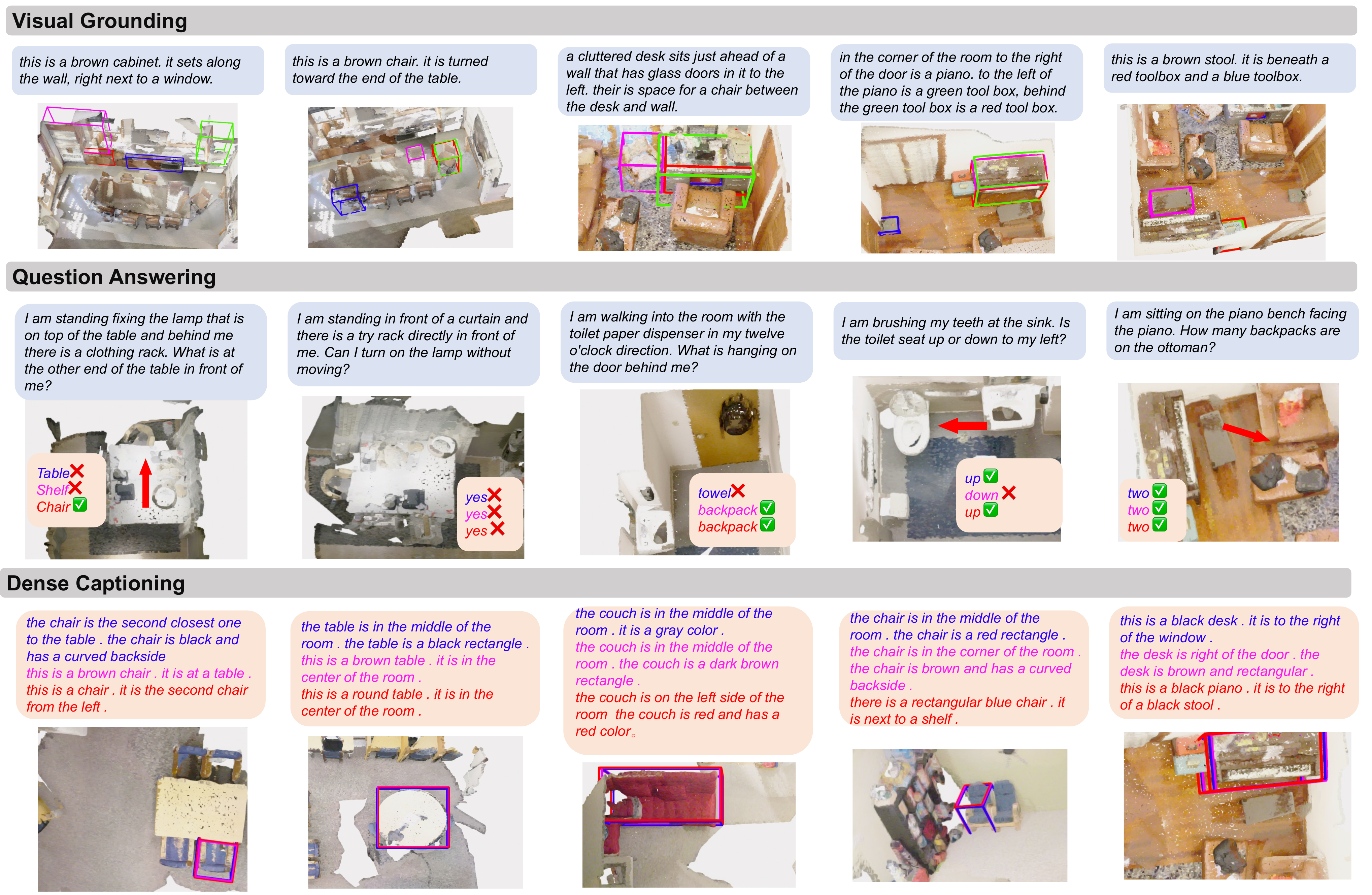}
    \caption{Comparison between different features. The \textcolor{blue}{blue} color represents the results obtained using only the voxel feature. The \textcolor{magenta}{magenta} color signifies the results from using both voxel and point features. The results derived from using all features are denoted by the color \textcolor{red}{red}. The ground truth results are represented by the color \textcolor{green}{green}. }
    \label{fig:comparse-feat}
\end{figure}

\noindent \textbf{Effect of different features}
 As illustrated in \cref{fig:comparse-feat}, it is clear that integrating image and point features leads to a more refined understanding of the scene. In tasks such as visual grounding and question answering, having more features enables the model to better comprehend the semantics, such as objects like backpacks and cluttered desks, as well as spatial relationships like ``other end'', ``corner'', ``beneath'', and attributes like color and object state. In the case of dense captioning, incorporating image features is particularly beneficial for the model to describe more detailed information about the instance, including class, shape, and color.

\section{Full quantitative results}
We provide full quantitative results in \cref{tab:ScanRefer}, \cref{tab:ReferIt3D}, \cref{tab:Multi3DRefer}, \cref{tab:ScanQA}, \cref{tab:SQA3D}, \cref{tab:Scan2Cap}, including results of baselines and all variants of \model for abalation study. Specifically, \model (sg.) denotes the model trained on a single dataset rather than through unified joint training, \model (V) denotes the model with only voxel features, \model (V,P) denotes the model with only point and voxel features, \model (rm I,P), and \model (rm I) denotes training with all features but removing image, voxel or image only during inference. \model (2-layer) denotes the model with 2-layer decoder, \model (6-layer) denotes the model with 6-layer decoder, \model (par.) denotes the model with parallel structure, and \model (seq.) denotes the model with sequential structure.

\begin{table*}[ht]
\caption{Grounding accuracy (\%) on ScanRefer with detected object masks. ``Det.'' represents the 3D object detection module used in the model. ``VN'' stands for VoteNet ~\cite{votenet}, ``PG'' for PointGroup ~\cite{pointgroup}, and M3D for Mask3D ~\cite{mask3d}, while ``Opt.'' denotes jointly optimizing the object detector on ScanRefer.}
\centering
\begin{tabular}{lccccccc}
\toprule
 \multirow{2}{*}{ Method } & \multicolumn{1}{c}{ Det. } & \multicolumn{2}{c}{ Unique } & \multicolumn{2}{c}{ Multiple } & \multicolumn{2}{c}{ Overall } \\
& & acc@0.25 & acc@0.5 & acc@0.25 & acc@0.5 & acc@0.25 & acc@0.5 \\
\midrule 3DVG-Trans ~\cite{3dvg} & Opt. & 81.9 & 60.6 & 39.3 & 28.4 & 47.6 & 34.7 \\
3D-SPS \cite{3dsps} & Opt. &  84.1  & 66.7 & 40.3 & 29.8 & 48.8 & 37.0 \\
3DJCG ~\cite{3djcg} & Opt. & 83.5 & 64.3 & 41.4 & 30.8 & 49.6 & 37.3 \\ SAT ~\cite{sat} & VN & 73.2 & 50.8 & 37.6 & 25.2 & 44.5 & 30.1 \\
MVT ~\cite{mvt3d} & PG & 77.7 & 66.5 & 31.9 & 25.3 & 40.8 & 33.3 \\
ViL3DRel ~\cite{vil3dref} & PG & 81.6 & 68.6 & 40.3 & 30.7 & 47.9 & 37.7 \\ 3D-VisTA ~\cite{3d-vista} & PG & 77.0 & 67.9 & 37.9 & 30.4 & 45.2 & 37.3 \\
3D-VisTA ~\cite{3d-vista} & M3D & 81.6 &  75.1  &  43.7  &  39.1  &  50.6  &  45.8  \\
\midrule 
\model (\textit{sg.}) & Opt. & 85.2 & 76.6 & 46.8 & 42.0 & 52.8 & 47.4 \\
\model & Opt. & 86.7 & 78.3 & 51.5 & 46.2 & 57.0 & 51.2 \\
\model (2-layer) & Opt. & 86.6 & 78.2 & 50.6 & 45.2 & 56.1 & 50.2 \\
\model (6-layer) & Opt. & 85.0 & 77.1 & 51.2 & 45.9 & 56.4 & 50.7 \\
\model (V) & Opt. & 82.9 & 75.1 & 45.9 & 40.9 & 51.6 & 46.1 \\
\model (rm I,P) & Opt. & 84.0 & 76.4 & 46.8 & 41.7 & 52.6 & 47.1 \\
\model (V,P) & Opt. & 84.4 & 76.1 & 49.7 & 44.4 & 55.0 & 49.2 \\
\model (rm I) & Opt. & 86.2 & 77.9 & 49.1 & 44.3 & 54.8 & 49.4 \\
\model (par.) & Opt. & 85.8 & 77.5 & 50.9 & 45.6 & 56.3 & 50.5 \\
\model (seq.) & Opt. & 83.7 & 76.1 & 46.8 & 41.9 & 52.5 & 47.1 \\
\bottomrule
\end{tabular}
\label{tab:ScanRefer}
\end{table*}

\begin{table*}[ht]
\caption{Grounding accuracy (\%) on Nr3D and Sr3D with ground-truth object masks.}
\centering
\begin{tabular}{lccccc|ccccc}
\toprule & \multicolumn{5}{c|}{ Nr3D } & \multicolumn{5}{c}{ Sr3D } \\
\cline { 2 - 11 } Method & Overall & Easy & Hard & \begin{tabular}{c} 
View \\
Dep
\end{tabular} & \begin{tabular}{c} 
View \\
Indep
\end{tabular} & Overall & Easy & Hard & \begin{tabular}{c} 
View \\
Dep
\end{tabular} & \begin{tabular}{c} 
View \\
Indep
\end{tabular} \\
\midrule 3DVG-Trans ~\cite{3dvg} & 40.8 & 48.5 & 34.8 & 34.8 & 43.7 & 51.4 & 54.2 & 44.9 & 44.6 & 51.7 \\
TransRefer3D ~\cite{transrefer3d} & 48.0 & 56.7 & 39.6 & 42.5 & 50.7 & 57.4 & 60.5 & 50.2 & 49.9 & 57.7 \\
LAR ~\cite{lar} & 48.9 & 58.4 & 42.3 & 47.4 & 52.1 & 59.4 & 63.0 & 51.2 & 50.0 & 59.1 \\
SAT ~\cite{sat} & 56.5 & 64.9 & 48.4 & 54.4 & 57.6 & 57.9 & 61.2 & 50.0 & 49.2 & 58.3 \\
3D-SPS ~\cite{3dsps} & 51.5 & 58.1 & 45.1 & 48.0 & 53.2 & 62.6 & 56.2 & 65.4 & 49.2 & 63.2 \\
MVT ~\cite{mvt3d} & 59.5 & 67.4 & 52.7 & 59.1 & 60.3 & 64.5 & 66.9 & 58.8 & 58.4 & 64.7 \\
ViL3DRel ~\cite{vil3dref} &  64.4  & 70.2 &  57.4  &  62.0  & 64.5 & 72.8 & 74.9 & 67.9 &  63.8  & 73.2 \\
3D-VisTA ~\cite{3d-vista} & 64.2 &  72.1  & 56.7 & 61.5 &  65.1  &  76.4  &  78.8  &  71.3  & 58.9 &  77.3  \\
\midrule 
\model (\textit{sg.}) & 64.9 & 73.3 & 56.7 & 60.7 & 67.0 & 75.6 & 78.8 & 68.2 & 51.5 & 76.7 \\
\model & 66.7 & 75.0 & 58.7 & 62.8 & 68.6 & 79.7 & 82.7 & 72.8 & 62.9 & 80.5 \\
\model (2-layer) & 64.2 & 72.7 & 56.3 & 57.2 & 67.8 & 78.2 & 81.3 & 70.9 & 62.1 & 78.9 \\
\model (6-layer) & 66.5 & 74.1 & 59.4 & 60.9 & 69.3 & 80.0 & 82.6 & 73.9 & 65.8 & 80.7 \\
\model (V) & 58.8 & 66.7 & 51.4 & 55.2 & 60.7 & 71.4 & 73.5 & 66.5 & 56.9 & 72.1 \\
\model (rm I,P) & 60.3 & 69.0 & 52.0 & 56.2 & 62.4 & 71.9 & 75.0 & 64.9 & 55.5 & 72.7 \\
\model (V,P) & 64.4 & 73.3 & 56.0 & 61.4 & 65.9 & 77.5 & 80.7 & 70.0 & 61.0 & 78.3 \\
\model (rm I) & 63.7 & 71.8 & 56.1 & 59.1 & 66.0 & 77.3 & 80.4 & 70.1 & 61.7 & 78.0 \\
\model (par.) & 65.9 & 74.2 & 58.1 & 62.5 & 67.6 & 78.7 & 81.8 & 71.5 & 65.7 & 79.3 \\
\model (seq.) & 56.1 & 63.8 & 48.9 & 47.6 & 60.4 & 70.8 & 73.6 & 64.3 & 56.1 & 71.5 \\
\bottomrule
\end{tabular}
\label{tab:ReferIt3D}
\end{table*}

\begin{table*}[ht]
\caption{Grounding accuracy (\%) on Multi3DRefer. Results of 3DVG-Trans+, D3Net and 3DJCG are provided by ~\cite{multi3drefer}.}
\centering
\begin{tabular}{lcccccc}
\toprule \multirow{2}{*}{ Method } & \multicolumn{6}{c}{ F1@0.5 (Pred boxes) } \\
& ZT w/o D & ZT w/D & ST w/o D & ST w/D & MT & All \\
\midrule 3DVG-Trans+ ~\cite{3dvg} & 87.1 & 45.8 & 27.5 & 16.7 & 26.5 & 25.5 \\
D3Net (Grounding) ~\cite{d3net} & 81.6 & 32.5 & 38.6 & 23.3 & 35.0 & 32.2 \\
3DJCG (Grounding) ~\cite{3djcg} & 94.1 & 66.9 & 26.0 & 16.7 & 26.2 & 26.6 \\
M3DRef-CLIP ~\cite{multi3drefer} & 81.8 & 39.4 & 47.8 & 30.6 & 37.9 & 38.4 \\
\midrule 
\model (\textit{sg.}) & 87.1 & 61.1 & 66.2 & 40.5 & 41.7 & 48.6 \\
\model & 85.4 & 57.7 & 68.5 & 43.6 & 40.9 & 50.1 \\
\model (2-layer) & 85.2 & 56.9 & 66.3 & 41.4 & 39.0 & 48.1 \\
\model (6-layer) & 86.5 & 58.5 & 67.4 & 44.0 & 41.6 & 50.3 \\
\model (V) & 84.1 & 55.1 & 63.0 & 36.8 & 37.5 & 44.8 \\
\model (rm I,P) & 84.1 & 57.7 & 62.9 & 38.2 & 38.1 & 45.7 \\
\model (V,P) & 82.4 & 54.6 & 66.0 & 40.9 & 39.4 & 47.7 \\
\model (rm I) & 85.6 & 55.4 & 66.9 & 40.8 & 39.8 & 48.1 \\
\model (par.) & 87.8 & 58.5 & 66.8 & 42.4 & 41.1 & 49.4 \\
\model (seq.) & 84.5 & 58.8 & 62.3 & 34.5 & 35.5 & 43.2 \\
\bottomrule
\end{tabular}
\label{tab:Multi3DRefer}
\end{table*}

\begin{table*}[ht]
\caption{Answer accuracy on ScanQA. Each entry denotes ``test w/ object'' / ``test w/o object''.}
\centering
\begin{tabular}{lccccc}
\toprule Method & EM@1 & BLEU-1 & ROUGE & METEOR & CIDEr \\
\midrule Image+MCAN ~\cite{scanqa} & 22.3 / 20.8 & 26.7 / 26.3 & 31.3 / 29.2 & 12.1 / 11.5 & 60.4 / 55.6 \\ ScanRefer+MCAN ~\cite{scanqa} & 20.6 / 19.0 & 27.9 / 27.0 & 30.7 / 28.6 & 12.0 / 11.4 & 57.4 / 53.4 \\ ScanQA ~\cite{scanqa} & 23.5 / 20.9 & 31.6 / 30.7 & 34.3 / 31.1 & 13.6 / 12.6 & 67.3 / 60.2 \\ 3D-VisTA ~\cite{3d-vista} & 27.0 / 23.0 & 34.4 / 30.2 & 38.6 / 32.8 & 15.2 / 12.9 & 76.6 / 62.6 \\ \midrule 
\model (\textit{sg.})       & 18.9 / 16.1 & 34.7 / 30.5  & 35.6 / 30.4 & 14.5 / 12.1 & 69.3 / 56.0 \\
\model              & 26.1 / 20.0 & 43.0 / 36.1& 42.9 / 34.0 & 17.8 / 13.9 & 87.8 / 65.2 \\
\bottomrule
\end{tabular}
\label{tab:ScanQA}
\end{table*}

\begin{table*}[ht]
\caption{Answer accuracy on SQA3D under question types.}
\centering
\begin{tabular}{lccccccc}
\toprule
\multirow{2}{*}{Method} & \multicolumn{6}{c}{Test set} & \multirow{2}{*}{Avg.} \\ 
\cline{2-7}
& What & Is & How & Can & Which & Other & \\
\midrule GPT-3 ~\cite{gpt-3} & 39.7 & 46.0 & 40.5 & 45.6 & 36.1 & 38.4 & 41.0 \\
ClipBERT ~\cite{clipbert} & 30.2 & 60.1 & 38.7 & 63.3 & 42.5 & 42.7 & 43.3 \\
SQA3D(w/o s) ~\cite{sqa3d} & 28.6 & 65.0 & 47.3 & 66.3 & 43.9 & 42.9 & 45.3 \\
SQA3D ~\cite{sqa3d} & 31.6 & 63.8 & 46.0 & 69.5 & 43.9 & 45.3 & 46.6 \\
3D-VisTA ~\cite{3d-vista} & 34.8 & 63.3 & 45.4 & 69.8 & 47.2 & 48.1 & 48.5 \\
\midrule 
\model (\textit{sg.}) & 35.6 & 62.7 & 45.2 & 66.3 & 43.3 & 43.3 & 46.8 \\
\model & 37.1 & 61.4 & 44.5 & 61.0 & 47.0 & 45.1 & 47.1 \\
\model (2-layer) & 37.7 & 62.1 & 41.4 & 62.9 & 36.2 & 46.2 & 46.3 \\
\model (6-layer) & 36.2 & 62.2 & 42.0 & 64.6 & 40.4 & 42.5 & 46.0 \\
\model (V) & 32.3 & 59.2 & 40.1 & 61.5 & 43.2 & 41.3 & 43.7 \\
\model (rm I,P) & 31.5 & 60.3 & 41.1 & 61.8 & 42.7 & 44.5 & 44.2 \\
\model (V,P) & 33.5 & 59.7 & 40.3 & 66.7 & 41.8 & 46.5 & 45.4 \\
\model (rm I) & 35.4 & 61.0 & 42.4 & 62.3 & 43.5 & 43.9 & 45.8 \\
\model (par.) & 37.7 & 60.9 & 42.0 & 64.1 & 42.1 & 41.4 & 46.2 \\
\model (seq.) & 33.6 & 58.9 & 43.3 & 65.5 & 42.6 & 43.7 & 45.2 \\
\bottomrule
\end{tabular}
\label{tab:SQA3D}
\end{table*}

\begin{table*}[ht]
\caption{Captioning results on Scan2Cap dataset. ``C'' stands for ``CIDEr'', ``B-4'' for ``BLEU-4'', ``M'' for ``METEOR'', and ``R'' for ``ROUGE'', respectively. ``@0.25'' and ``@0.5'' represent the overlap ratios between the predicted boxes and ground truth boxes.}
\centering
\begin{tabular}{lcccc|cccc}
\toprule \multirow{2}{*}{ Method } & \multicolumn{4}{c|}{ @0.25 } & \multicolumn{4}{c}{ @0.5 } \\
&  C  &  B-4  &  M  &  R  &  C  &  B-4  &  M  &  R  \\
\midrule Scan2Cap ~\cite{scan2cap} & 53.7 & 34.3 & 26.1 & 55.0 & 35.2 & 22.4 & 21.4 & 43.5 \\
3DJCG ~\cite{3djcg} & 60.9 &  39.7  & 27.5 &  59.0  & 47.7 & 31.5 & 24.3 & 51.8 \\
3D-VisTA ~\cite{3d-vista} &  71.0  & 36.5 &  28.4  & 57.6 &  66.9  &  34.0  &  27.1  & 54.3 \\
\midrule 
\model (\textit{sg.}) & 81.5 & 37.5 & 30.4 & 60.6 & 75.6 & 34.5 & 28.6 & 57.1 \\
\model & 87.1 & 39.2 & 30.9 & 61.5 & 80.3 & 36.0 & 29.1 & 57.9 \\
\model (2-layer) & 86.4 & 38.8 & 30.7 & 61.0 & 79.8 & 35.5 & 28.8 & 57.3 \\
\model (6-layer) & 83.2 & 36.2 & 30.0 & 59.5 & 76.7 & 33.1 & 28.1 & 56.0 \\
\model (V) & 73.3 & 33.8 & 28.8 & 58.5 & 67.8 & 31.0 & 27.1 & 54.9 \\
\model (rm I,P) & 73.2 & 34.2 & 29.1 & 59.3 & 68.1 & 31.5 & 27.4 & 55.9 \\
\model (V,P) & 80.2 & 37.7 & 30.2 & 60.9 & 74.6 & 34.8 & 28.4 & 57.5 \\
\model (rm I) & 80.4 & 37.8 & 30.1 & 60.7 & 74.7 & 34.7 & 28.4 & 57.2 \\
\model (par.) & 84.9 & 39.1 & 30.5 & 61.2 & 78.3 & 35.7 & 28.6 & 57.5 \\
\model (seq.) & 66.1 & 29.1 & 27.9 & 56.1 & 61.4 & 27.0 & 26.3 & 53.0 \\
\bottomrule
\end{tabular}
\label{tab:Scan2Cap}
\end{table*}

\end{document}